\documentclass[sn-mathphys,iicol,pdflatex]{sn-jnl}% Math and Physical Sciences Reference Style
\graphicspath{{./figures/}}
\usepackage{enumitem}
\usepackage{ulem}
\usepackage{amssymb}
\usepackage{amsfonts}
\usepackage{amsmath}
\usepackage{mathrsfs}
\usepackage{float}
\makeatletter
\newcommand{\cb}[1]{{\color{uniSblue}{#1}}}
\newcommand{\clb}[1]{{\color{uniSlightblue}{#1}}}
\newcommand{\cg}[1]{{\color{uniSgreen}{#1}}}
\newcommand{\co}[1]{{\color{uniSorange}{#1}}}
\newcommand{\relmse}[0]{{rel.~$\sqrt{\text{MSE}}$}  }
\newcommand{\figref}[1]{Fig.~\ref{#1}}

\newcommand{\tabref}[1]{tab.~\ref{#1}}

\newcommand{\secref}[1]{section~\ref{#1}}

\newcommand{\equref}[1]{equation~(\ref{#1})}

\usepackage{qrcode}
\usepackage{stmaryrd}

\newcommand{\sM}{\begin{array}{ccccccccc}}
\newcommand{\eM}{\end{array}}

\newcommand{\lb}{\left(}
\newcommand{\rb}{\right)}

\newcommand{\sv}{\lb\begin{array}{ccccccccccccccccc}}
\newcommand{\sV}{\begin{bmatrix}}
\newcommand{\eV}{\end{bmatrix}}
\newcommand{\ev}{\end{array}\rb}

\newcommand{\fempty}[1]{{}}

\newcommand{\sty}[1]{{\boldsymbol{#1}}}
\newcommand{\styy}[1]{{\mathbb{#1}}}

\newcommand{\fx}{\sty{ x}}

\newcommand{\ffA}{\styy{ A}}

\newcommand{\ffN}{\styy{ N}}

\newcommand{\cF}{{\cal F}}

\newcommand{\cN}{{\cal N}}

\newcommand{\WT}[1]{\widetilde{#1}}
\newcommand{\WH}[1]{\widehat{#1}}
\newcommand{\ol}[1]{\overline{#1}}

\newcommand{\ul}[1]{\underline{#1}}

\newcommand{\ull}[1]{\ul{\ul{#1}}}

\makeatletter
\definecolor{Sblueaa}{cmyk}{1,0.6,0,0}
\definecolor{Sbluea}{cmyk}{1,0.4,0,0}
\definecolor{Sblueb}{cmyk}{0.7,0.2,0,0}
\definecolor{Sbluec}{cmyk}{0.5,0.1,0,0}
\definecolor{Sblued}{cmyk}{0.3,0.05,0,0}
\definecolor{Sbluee}{cmyk}{0.15,0.04,0,0}
\definecolor{Svbluea}{cmyk}{0.9,0.6,0,0}
\definecolor{Svblueb}{cmyk}{0.68,0.4,0,0}
\definecolor{Svbluec}{cmyk}{0.45,0.26,0,0}
\definecolor{Svblued}{cmyk}{0.27,0.12,0,0}
\definecolor{Sblacka}{cmyk}{0.5,0.2,0.2,0.85}
\definecolor{Sblackb}{cmyk}{0.35,0.14,0.14,0.6}
\definecolor{Sblackc}{cmyk}{0.25,0.1,0.1,0.43}
\definecolor{Sblackd}{cmyk}{0.15,0.06,0.06,0.26}
\definecolor{Sblacke}{rgb}{0.827451,0.8509804,0.8627451}
\definecolor{Sred100}{HTML}{EE1C25}
\definecolor{Sorange100}{HTML}{F36F23}
\definecolor{Syellow100}{HTML}{FFDD00}
\definecolor{Spetrol}{HTML}{00AAAD}
\definecolor{Sgreen100}{HTML}{8DC63F}
\definecolor{Spink100}{HTML}{EC008D}
\definecolor{Spurple100}{HTML}{812A91}
\definecolor{Syellow}{cmyk}{0,0.1,1,0}
\definecolor{Sorange}{cmyk}{0,0.7,1,0}
\definecolor{Sred}{cmyk}{0,1,1,0}
\definecolor{Spink}{cmyk}{0,1,0,0}
\definecolor{Spurple}{cmyk}{0.6,1,0,0}
\definecolor{Scyan}{cmyk}{1,0,0.4,0}
\definecolor{Sgreen}{cmyk}{0.5,0,1,0}
\definecolor{Sgreen}{cmyk}{0.5,0,1,0}
\definecolor{uniSgreen}{HTML}{93FF00}
\definecolor{uniSred}{HTML}{FF000B}
\definecolor{uniSpink}{HTML}{FF0098}
\definecolor{uniSorange}{HTML}{FF5D00}
\definecolor{uniScyan}{HTML}{00FBFF}

\colorlet{Sblackf}{Sblacke!70!white}
\colorlet{Sdgreen}{Sgreen!50!black}
\colorlet{Slred}{Sred!40!white}

\definecolor{uniSgray}{RGB}{62, 68, 76}
\colorlet{USredgray}{red!70!uniSgray}
\colorlet{uniSredgray}{red!70!uniSgray}
\colorlet{uniSgray90}{uniSgray!90!white}
\colorlet{uniSgray80}{uniSgray!80!white}
\colorlet{uniSgray70}{uniSgray!70!white}
\colorlet{uniSgray60}{uniSgray!60!white}
\colorlet{uniSgray50}{uniSgray!50!white}
\colorlet{uniSgray40}{uniSgray!40!white}
\colorlet{uniSgray30}{uniSgray!30!white}
\colorlet{uniSgray20}{uniSgray!20!white}
\colorlet{uniSgray10}{uniSgray!10!white}
\definecolor{Scyanlight}{rgb}{ 0.53333,0.87059,0.87451}
\definecolor{uniSblue}{HTML}{004191}
\colorlet{uniSblue80}{uniSblue!80!white}
\colorlet{uniSblue60}{uniSblue!60!white}
\colorlet{uniSblue40}{uniSblue!40!white}
\definecolor{uniSlightblue}{HTML}{00BEFF}
\colorlet{uniSlblue80}{uniSlightblue!80!white}
\colorlet{uniSlblue60}{uniSlightblue!60!white}
\colorlet{uniSlblue40}{uniSlightblue!40!white}
\definecolor{FFgreen}{rgb}{0.635,0.8275,0.1255}
\definecolor{FFdgreen}{rgb}{0.45,0.61,0.09}

\newcommand{\Stilde}{\raise.17ex\hbox{$\scriptstyle\sim$}}

\makeatother

\usepackage[absolute,overlay]{textpos}
\TPGrid{100}{100}
\usepackage{hyperref}
\makeatother
\floatname{algorithm}{Procedure}

\jyear{2022}%

\theoremstyle{thmstyleone}%
\theoremstyle{thmstyletwo}%

\theoremstyle{thmstylethree}%

\begin{document}

\title[Hybrid machine-learned microstructure homogenization]{Hybrid machine-learned homogenization: Bayesian data mining and convolutional neural networks}


\author*[1]{\fnm{Li\ss ner} \sur{Julian}}\email{lissner@mib.uni-stuttgart.de}

\author[1]{\fnm{Fritzen} \sur{Felix}}\email{fritzen@simtech.uni-stuttgart.de}

\affil*[1]{\orgdiv{Data Analytis in Engineering}, \orgname{University of Stuttgart}, \orgaddress{\street{Universit\"atsstra\ss e 32}, \city{Stuttgart}, \postcode{70569}, \state{Baden W\"urttemberg}, \country{Germany}}}


\abstract{Beyond the generally deployed features for microstructure property prediction this study aims to improve the machine learned prediction by developing novel feature descriptors. Therefore, Bayesian infused data mining is conducted to acquire samples containing characteristics inexplicable to the current feature set, and suitable feature descriptors to describe these characteristics are proposed. The iterative development of feature descriptors resulted in 37 novel features, being able to reduce the prediction error by roughly one third. To further improve the predictive model, convolutional neural networks (Conv Nets) are deployed to generate auxiliary features in a supervised machine learning manner. The Conv Nets were able to outperform the feature based approach. A key ingredient for that is a newly proposed data augmentation scheme and the development of so-called deep inception modules. A combination of the feature based approach and the convolutional neural network leads to a hybrid neural network: A parallel deployment of the both neural network archetypes in a single model achieved a relative rooted mean squared error below 1\%, more than halving the error compared to prior models operating on the same data. The hybrid neural network was found powerful enough to be extended to predict variable material parameters, from a low to high phase contrast, while allowing for arbitrary microstructure geometry at the same time. }




\maketitle

\section{Introduction}

High performance materials are of great interest to the industry due to their capabilities and scope of application, e.g., in aerospace applications or for batteries, even though their development and manufacturing process is a highly challenging task. The tailoring of specific materials can be conducted by, e.g., tuning its microstructure to optimize its properties given specific requirements. The development process can be significantly boosted by replacing experimental tests through simulations, which are carried out by taking the microscopic geometric information of the materials into account \citep{miehe2002,beyerlein2008dislocation}. Some simulation methods are able to directly operate on the 3D image representation of the microstructure, e.g., obtained from a computed tomography (CT) scan, recently achieving improvements with respect to computational speed \cite{keshav2022fft}. The high resolution of the image representation renders even such efficient methods infeasible to evaluate in a many query context by investigating numerous (e.g., in-silico generated) microstructured materials, when optimizing for specific material behaviour.

\begin{figure*}[tp!]
   \centering\includegraphics[width=\textwidth]{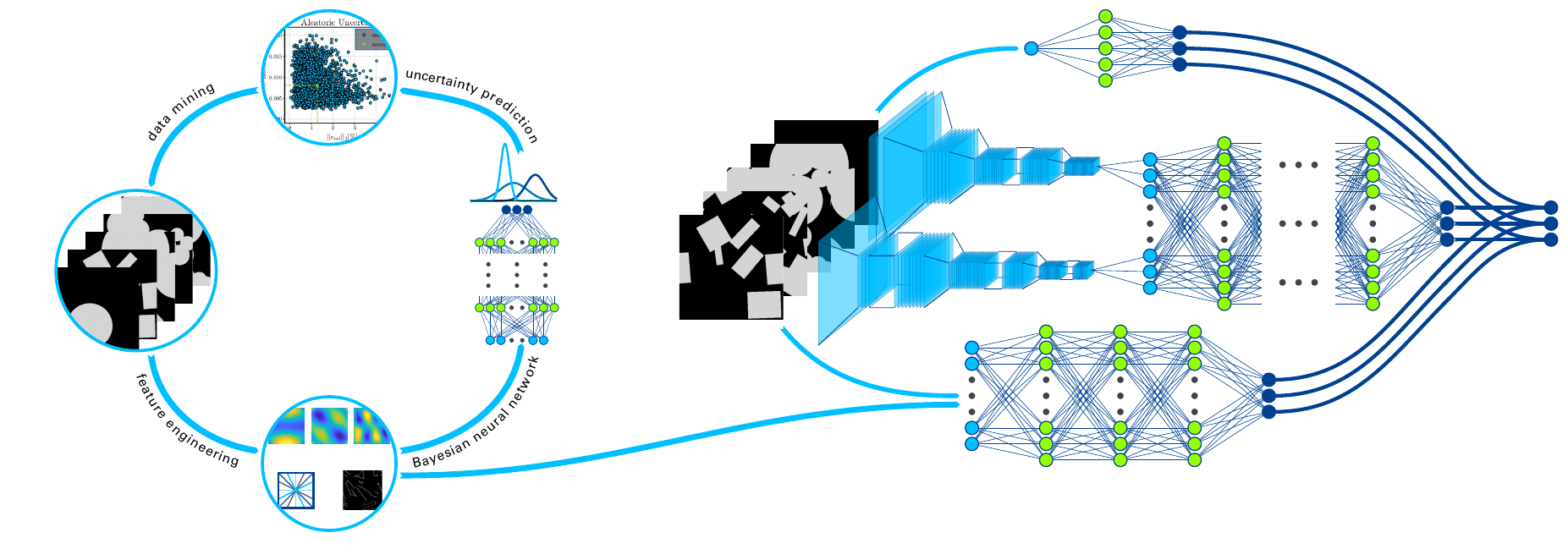}
  \caption{The general outline of this paper is presented in figure format. On the left, the iterative approach of Bayesian assisted data mining is displayed, and the newly developed features thereof are used in a \textit{hybrid neural network} (on the right). The backbone of the hybrid neural network are \textit{deep inception modules} in parralel to a bypass using the volume fraction and the feature regressor. }\label{fig.approach} 
\end{figure*}

Machine learning is a suitable tool to further reduce the computational cost for  material development: It can discover complex relationships by studying the available data and replacing the costly simulations with computationally affordable operations. Various machine learning algorithms are regularly deployed in two actively researched fields of microstructure modeling, namely microstructure reconstruction and microstructure property prediction. In microstructure reconstruction/synthesis, the topology of an original microstructured image is adjusted to optimize for selected material behaviour under constraints \citep{kumar2016markov,cang2018improving,seibert2022descriptor}, tailoring for specific microstructures to serve a particular purpose. The field of microstructure property prediction is often applied to a broader spectrum of microstructured materials \cite{lissner2019data,ford2021machine}. It targets the efficient and accurate prediction of the behaviour of inhomogeneous materials via, e.g., artificial neural networks (ANN).

The present paper falls into a subclass of the latter category, specifically microstructure homogenization, which directly predicts effective material properties from given microstructural image data. Current state of the art methods often deploy a blend of different unsupervised and supervised machine learning methods, where the Principal Component Analysis (PCA) is used in combination of the 2-Point Correlation Function (2PCF) \citep{brough2017materials}, using the principal scores as input for a supervised machine learned regressor to conduct the microstructure property linkage, ranging from polynomial regression \citep{marshall2021autonomous} to artificial neural networks \citep{lissner2019data,farizhandi2022processing}. Lately, the usage of convolutional neural networks (Conv Nets) has gained popularity in both outlined research fields \citep{lubbers2017inferring,gayon2020pores}. Conv Nets are used to directly predict the effective material response \citep{liu2022correlation}, or even in combination with the PCA where the reduced representation of the target values is used to predict stress strain relationships \citep{yang2020prediction}. Conv Nets have the advantage over classical regressors in the sense that they are better suited to an extension for the prediction of full field solution in voxel (3D pixels) representation, since the data is often given in image representation, for the microstructured material as well as the full field response.

The previous study of the authors \citep{lissner2019data} has deployed a method akin to the PCA and used the derived features in various state of the art machine learned regressors, finding that the accuracy of the approach is limited, which is confirmed in different studies, e.g., \citep{ford2021machine, fernandez2020generation}. Improvements beyond the 2PCF have been attempted by considering partial higher order correlation functions, e.g., \citep{fast2011formulation}. However, the feature identification and computation becomes increasingly costly while yielding rather limited improvements with respect to accuracy. In the search for a better prediction, the idea of auxiliary features came to mind, which led us to data mining \citep{mikut2011data}. This constitutes the first block of this work (\secref{sec.Bayesian}): We systematically categorize the underlying data by using Bayesian neural networks \citep{tipping2003bayesian}. We evaluate the aleatoric uncertainty which can hint at a lack of feature knowledge for certain samples. These samples can then be examined systematically in order to engineer additional features while being machine-guided during the feature engineering process. The thereby identified recurrent characteristics across multiple samples were then quantified by novel feature descriptors, while keeping computational efficiency and physical interpretability in consideration.

The second block of this study considers Conv Nets (\secref{sec.conv_net}), which are able to derive machine learned features via supervised learning. Further improvements with respect to Conv Nets are found by building upon the so called inception modules \citep{szegedy2015going} and, therefrom, developing a \textit{deep inception module}. The latter is explicitely designed to be able to capture features at different length scales within a single microstructural image.
Ultimately, we combine the handcrafted, machine-guided features with features derived from Conv Nets into a \textit{hybrid neural network}.  
The proposed model more than halved the prediction error compared to previous studies \citep{lissner2019data}. This hybrid model is further extended to predict the effective material properties for variable material parameters, allowing for variable phase contrast ranging from $\frac12$ up to $\frac1{100}$ while considering variable microstructure material characteristics, e.g., with the volume fraction ranging from 20-80\%. 

The data \cite{darus1151} used for development, training and validation purposes, as well as the \texttt{python} code \cite{codebase} are made freely available.

\section{Data and Methods}\label{sec.methods}
\subsection{Data overview}\label{sec.data}

\begin{figure*}
\centering \includegraphics[height=0.20\textheight]{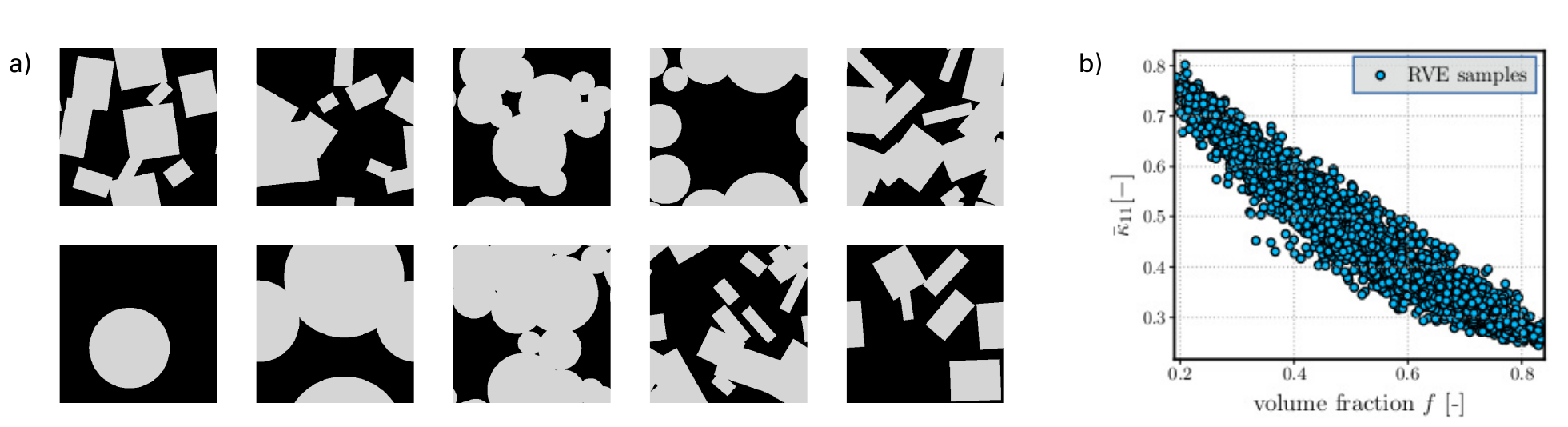}
\caption{The investigated data is shown, where a few RVE images of the training set are shown on the left. The gray foreground represents the inclusion phase, and the black background represents the matrix phase. On the right, one component of the target values is shown for all samples out of the training set.}\label{fig.data_overview}
\end{figure*}

%% remodel the sentence
In the machine learning focused manuscript, some emphasis has to be given to the data. This manuscript deals with the prediction of material properties from bi-phasic microstructured materials. The simplification of a Representative Volume Element (RVE) is introduced, where it is assumed that a single frame of the microstructure suffices to characterize the material behaviour of the macroscopic material \cite{torquato2013random} while using periodic boundary conditions. 
A compact overview of the data is given in \figref{fig.data_overview}, where some exemplary images of the microstructure are plotted on the left. These images, i.e., RVEs of the microstructured material serve as the input data to our algorithm and will be denoted $\ull{\rm RVE}$ in the following. The target values of the machine learning algorithm are the components of the effective heat conduction tensor $\ull{\bar\kappa}$. One Component of $\ull{\bar\kappa}$ is plotted in \figref{fig.data_overview} b), considering low conducting inclusions with a phase contrast of $R=5$. A more detailed explanation of the data is given in the appendix \ref{appendix.data}.
Generally, the symmetric effective heat conduction tensor will be given in de-dimensionalized Mandel notation 
\begin{equation}\label{eq.targets}
  \bar{\ul\kappa} = \begin{bmatrix} \bar\kappa_{11} \\ \bar\kappa_{22} \\ \sqrt2 \bar\kappa_{12} \end{bmatrix}\,.
\end{equation}

The objective of this paper is to find a machine learned model which is able to accurately predict the effective material properties of the presented microstructures, i.e.
\begin{equation}
  \bar{\ul\kappa} = f( {\rm RVE} )
\end{equation}
where the machine learned model $f(\cdot)$ directly operates on the image data of the microstructured material (${\rm RVE}$) or uses features $\ul x$ directly extracted from the image. The constraint to our algorithm is to solely rely on the given images of the RVEs, without any further information such as, e.g., the number or shape of inclusions. In this manuscript we first aim to find optimal features $\ul x$ and a suitable artificial neural network $f(\cdot)$ accurately mapping the complex image based relationship between features and the sought-after outputs.

\subsection{Bayesian Modeling via artificial neural networks}\label{sec.Bayesian}
\subsubsection{Aleatoric and epistemic uncertainty}

% motivation for uncertainty modeling
In supervised machine learning the training and testing data is always given with deterministic target values such that error measures are computable. However, during inference, reliable error measures are generally not accessible. Consequently, having access to a measure of model confidence indicating the likelihood for low or high prediction errors during inference is advantageous. 

Bayesian modeling approaches uncertainty quantification via \textit{aleatoric} and \textit{epistemic} uncertainties \citep{kendall2017uncertainties,depeweg2019modeling}.  
The epistemic uncertainty aims to represent the uncertainty arising based on the lack of knowledge about the mapping from input to output values. In the machine learning context it arises from the model being unable to recover the original function, if it even exists. 

The aleatoric uncertainty arises from the uncertainty in the system or data and aims to represent noise in the input data, i.e., different output values for similar or even identical input values.  
Considering, for instance \figref{fig.data_overview} b) from a machine learning viewpoint, the variation of the material response could be interpreted as noisy when considering only the volume fraction as input. These variations, however, are not due to noise, but arise due to different phenomena inexplicable when trying to predict the response exclusively by the materials volume fraction. Consequently, the aleatoric uncertainty can also be regarded as a measure of explainability in the data given the underlying (possibly incomplete) feature set.

Since our main interest lies in the development of novel features based on the examination of data, i.e., data mining, we employ the aleatoric uncertainty and use it to detect samples which contain characteristics not explicable by the current feature set. The aleatoric uncertainty is modeled using the \texttt{tensorflow probability distribution} library \citep{dillon2017tensorflow} via artificial neural networks by assuming that the priors over the predicted data follow a normal distribution. Practically speaking, the aleatoric uncertainty is modeled by the prediction of the mean and standard deviation $\hat\mu_i, \hat\sigma_i$ which define a normal distribution of the $i$th output component instead of a single deterministic value. The neural networks predicted normal distribution is given as
\begin{equation}
    f(\ul x) = \frac1{\sqrt{2\pi\hat\sigma_i^2}}\exp{\Big(\frac{\hat\mu_i - y_i}{2\hat\sigma_i^2}\Big)}\, = p( y_i\vert \ul x)\,,
\end{equation}
with the model $f$ using the feature vector $\ul x$ to predict the parameters $\hat\mu_i$ and $\hat\sigma_i$ for each component $y_i$ independently. 

Thus, the neural network has to predict twice the number of output variables when modeling the aleatoric uncertainty as described. The quantification of the aleatoric uncertainty during training enters through the loss $\Phi_i$ for target value $y_i$, which is derived by minimizing the Kullback-Leibler divergence \cite{depeweg2019modeling}:
\begin{equation}\label{eq.bnn_loss}
  \Phi_i\big(y_i, \cN(\hat\mu_i, \hat\sigma^2_{i} ) \big) = -\log\big( \frac12 p(y_i\lvert\cN (\hat\mu_i, \hat\sigma^2_{i}) + s) \big) ,
\end{equation}
where $p(y\lvert\cN (\hat\mu_i, \hat\sigma^2_{i}))$ is the probability of observing $y_i$ given the current distribution $\cN(\hat\mu_i, \hat\sigma^2_{i} )$. Since the logarithm quickly tends to $-\infty$ for very small values in $p \lb y\lvert\cN (\hat\mu_i, \hat\sigma^2_{i}) \rb$, the shift parameter $s>0$ has been introduced to stabilize the training. This greatly improved convergence behaviour of the model for $s=0.25$.  

For a more detailed explanation of the theoretical background and derivation of Bayesian neural networks the authors refer to the literature, e.g., \citep{goan2020Bayesian,tipping2003bayesian}, presenting a more generous outline to bayesian modeling and \cite{kabir2018neural} who reviews different modeling methods in uncertainty quantification. An example application using bayesian modeling is presented in \cite{kendall2017uncertainties}.
\subsubsection{Use of Bayesian neural networks in feature engineering}\label{sec.new_features}

As previously outlined, the aleatoric uncertainty can be used to indicate lack of feature knowledge that hinders more accurate predictions. For instance, the volume fraction alone is unable to consider particle shapes and their orientation distribution (\figref{fig.data_overview} b)), which is reflected in the aleatoric uncertainty. Investigating these samples of high aleatoric uncertainty, it can be seen that each of these samples contains one or several characteristics inexplicable by the current feature set. Exploiting this information, our data mining approach is guided by the Bayesian neural network (BNN). First, we filter out a subset of our data by considering only samples of high aleatoric uncertainty. With this subset at hand, we aim at finding suitable feature descriptors to quantify the apparent characteristics.

Since a major motivation for machine learning is about gaining computational efficiency, the selection of input features for the machine learned model should bear computational efficiency in mind. This motivates us to develop novel features which are either obtainable through computationally cheap operations, or derived via the convolution operation. 

The effect of the discrete convolution operation will be briefly motivated by considering the 1D Sobel operator $\ul k^{\rm S}$
\begin{equation}
    \ul k^{\rm S} = \frac12 \begin{bmatrix} -1,& 0,& 1 \end{bmatrix}
\end{equation}
which is applied to an arbitrary 1D signal $\ul s \in \styy R^{n_x}$ via a discrete convolution: for any admissible $i$, i.e., $1<i< n_x$
\begin{equation}\label{eq.conv}
  \begin{array}{rl}
    \ul f = \ul s \ast \ul k^{\rm S} \to f_i = &
    \sum\limits_{j=-1}^1 s_{i+j}\,\,k^{\rm S}_{j+2}\\[4mm]
    = & \frac12 (s_{i+1} - s_{i-1})\,\,,
  \end{array}
\end{equation}
where the gradient information is obtained in the feature map $\ul f$ (c.f. central difference).

Since the kernel can be arbitrarily chosen, if one was to replace the Sobel operator $\ul k^S$ with a normalized constant vector, one would recover the result of the moving average in the feature map $\ul f$.

\begin{figure}[h]
    \includegraphics[width=0.45\textwidth]{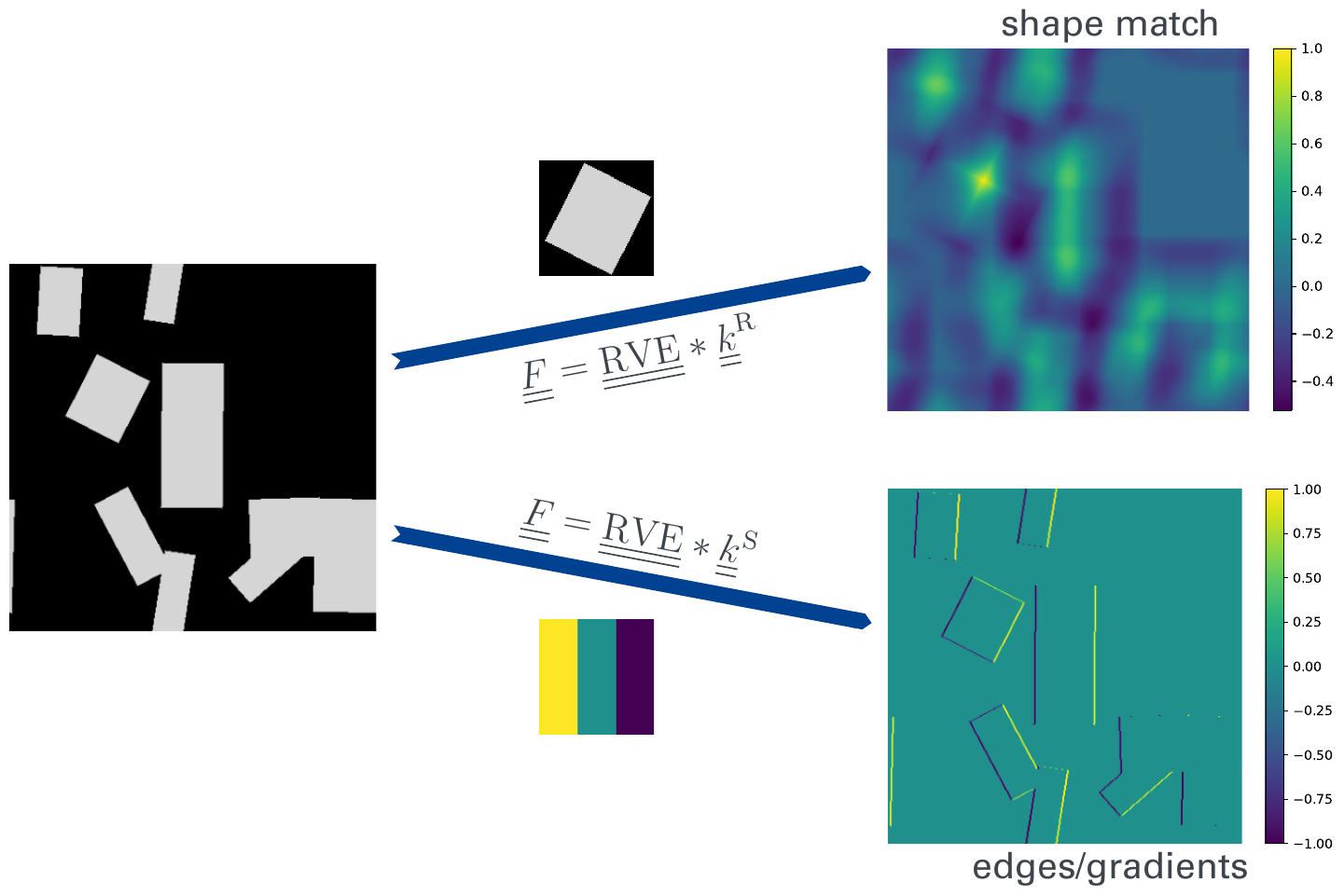}
    \caption{The convolution of one RVE with two different kernels is shown. The superscripts ${\rm S,R}$ of $\ull k$ represent the \textit{Sobel} filter and the searched \textit{rectangle}, respectively. The feature map $\ull F$ to the right differs significantly when applying different kernels to the same image.}\label{fig.kernel_examples}
\end{figure}

Analogously, the discrete convolution of a 2d signal, i.e., of an image, is conducted by adding a second dimension to the data and therefore a second summation to the operation (\equref{eq.conv}). A graphical overview of the convolution operation is given a little later in \secref{sec.conv_net} in the introduction of convolutional neural networks. Since the convolution is evaluated at each admissible index position with a double summation, the operation in itself becomes costly. The cost can be drastically reduced by conducting the convolution in Fourier space when following the convolutional theorem \cite{hunt1971matrix}, i.e.
\begin{equation}
  \ull F = \ull A \ast \ull k = \cF^{-1}\big( \,\cF(\ull A)\cdot\cF(\ull k) \big)\,.
\end{equation}
The discrete Fourier Transform induces periodicity, which is rather favorable in our case where the microstructure, i.e., the RVE, and the fields are periodic. 

\begin{figure*}%[bp]
    \centering\includegraphics[width=\textwidth]{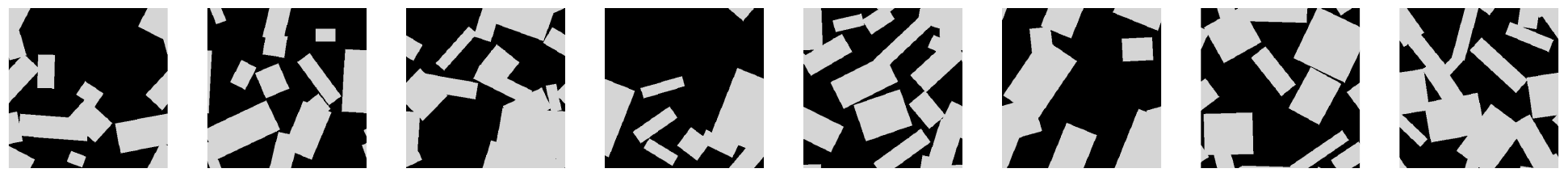}
  \caption{Iteration 1: A few samples which led to the highest relative prediction error and highest aleatoric uncertainties are shown. These samples were found by a model trained using only the reduced coefficients and were used for the first iteration of feature development.}\label{fig.special_samples_1}
\end{figure*}

Similar to the motivation in 1d is the resulting feature map $\ull F$ depends on the kernel $\ull k$, even when applying it to the same image $\ull A$. This makes the convolution operation exceedingly flexible for the quantification of characteristics. For instance, it could even be used to detect specific shapes by interpreting the feature map as a match indicator of the kernel in the image (\figref{fig.kernel_examples}). Thus, we develop feature descriptors by developing specific kernels and let the BNN guide us in the process.

\begin{figure}[h]
  \centering\includegraphics[height=5cm]{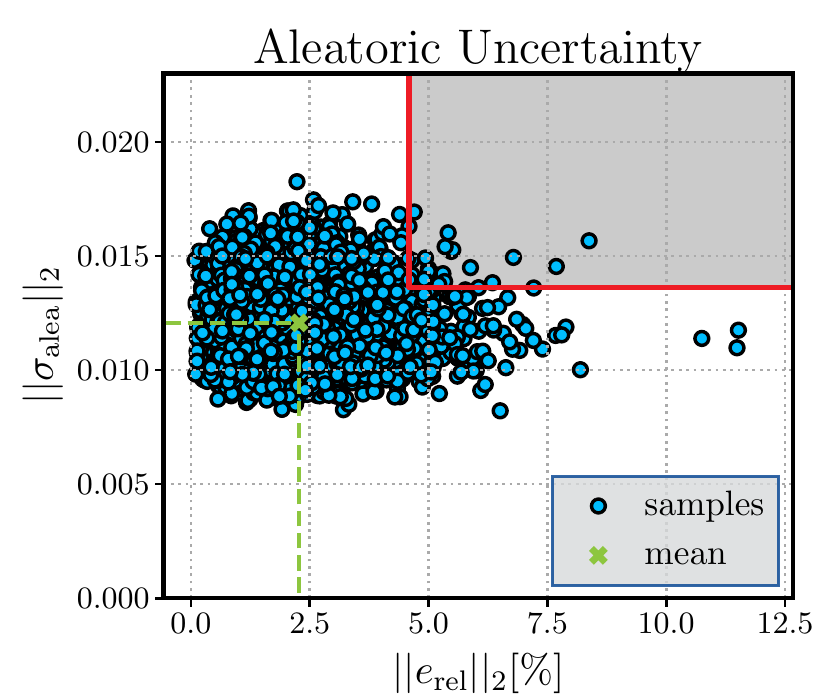}
  \caption{The relative prediction errors related to the aleatoric uncertainty are shown for the models prediction of the test set in the first iteration. Samples close to the top right corner (located inside the red box) are displayed in \figref{fig.special_samples_1} and used for feature engineering}\label{fig.prediction_scatter}
\end{figure}

We suggest the following general strategy: the entire set of currently available features is used to train a BNN. Thereafter, an additional data set, i.e., a test set consisting of unseen data, is predicted and the subset of samples of high prediction error and high aleatoric uncertainty (\figref{fig.prediction_scatter} inside the red box) are investigated to identify common phenomena by the data scientist/engineer. An efficient feature descriptor to quantify the patterns is proposed. The new features are subsequently added to the existing feature set, which are used to train a novel BNN. Therewith, the procedure was repeated and samples of high aleatoric uncertainty were investigated, postulating an iterative feature engineering approach. This is graphically illustrated in \figref{fig.approach} (left), and can be transferred to any application using human interpretable data.\\

\begin{figure*}%[h]
    \centering\includegraphics[width=\textwidth]{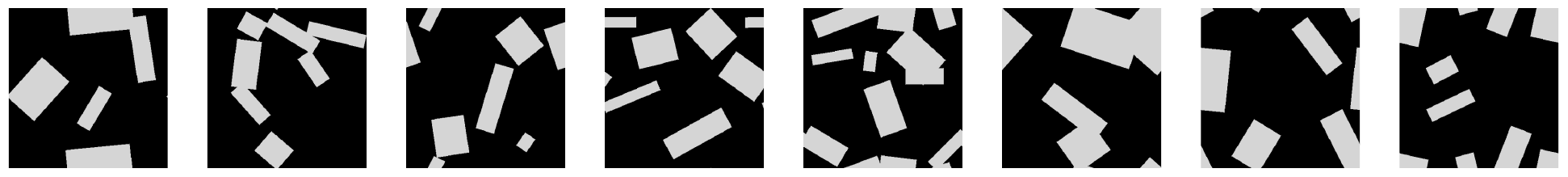}
  \caption{Iteration 2: A few samples which led to the highest relative prediction error and highest aleatoric uncertainties are shown. These samples were found by a model trained with the reduced coefficients and the band feature coeffcients. These images are studied in the second iteration of feature development.}\label{fig.special_samples_2}
\end{figure*}

The initial BNN was trained using the reduced coefficients of the two point correlation function (2PCF), as described in \citep{lissner2019data}. In order to isolate samples of high aleatoric uncertainty and high prediction error, we consider a relative error measure
\begin{equation}\label{eq.rel_norm}
  \|e_{\rm rel}\|_2 = \frac{ \| \ul{\bar\kappa} - \ul{\hat\kappa} \|_2 }{\| \ul{\bar\kappa} \|_2} 
\end{equation}
with the target value $\ul{\bar\kappa}$ and the prediction $\ul{\hat\kappa} \gets f(\ull{\rm RVE})$, which was taken as the mean $\hat\mu_{i}$ from the predicted normal distribution. The error-uncertainty relationship in the first iteration is displayed in \figref{fig.prediction_scatter}, where the entire test set was predicted. A few samples in the approximate region of the top right corner of \figref{fig.prediction_scatter} are shown in \figref{fig.special_samples_1}.

\begin{figure}[h]
  \centering \includegraphics[width=0.48\textwidth]{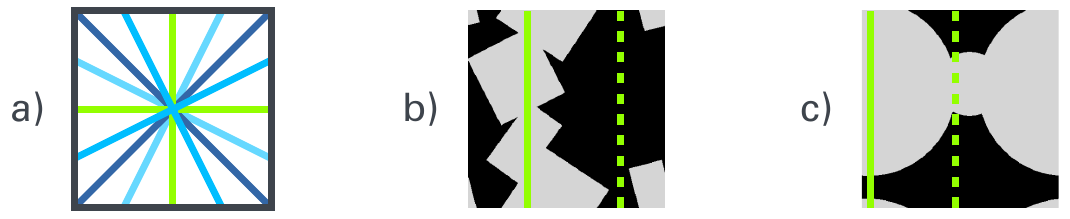}
  \caption{Each line on the left in a) denotes one scanned direction of the band feature detectors. The determination of the band feature in vertical direction is shown for two images in b) and c),  where the full lines measures the inclusion phase and the dashed line measures the matrix phase. The feature detector $\ull b_d$ in b) and c) are drawn in their approximate position for feature determination.}\label{fig.band_features}
\end{figure}

After close investigation of these samples, it was found that many of these samples of high aleatoric uncertainty share a diagonal structure, more strikingly do they contain connected regions or even percolation in certain directions. This led to the invention of the band features (\figref{fig.band_features}), which aim to detect linear connectivity. The band features have some resemblance with the lineal path function \citep{lu1992lineal}. The lineal path function has been previously adopted in multiple studies, e.g., being used as a feature \cite{kalidindi2011microstructure}, or as a descriptor to generate statistically similar RVEs \citep{scheunemann2015design,balzani2014construction}. The lineal path function computes the probability that a line segment of specific length under a certain angle lies fully within the inclusion phase. It is quite costly to compute for multiple angles and different lengths of the line segment.

The newly proposed band features are computed with a single convolution per direction $d$ as
\begin{equation}\label{eq.perco_incl}
  p^{\rm I}_d = \max(\ull b_d \ast \ull{\rm{RVE}}  )
\end{equation}
where $\ull b_d$ denotes the detection band in the indexed direction~$d$. Note that each detector $\ull b_d$ is normalized such that ${p^{\rm I}_d \in [0,1]}$. The maximum operation in \eqref{eq.perco_incl} virtually fixes the evaluation of the band feature on a single spot in the RVE, i.e., the location where the band feature detector is within the inclusion phase for the highest amount.

In addition to the linear connectivity found in the inclusion phase (\figref{fig.special_samples_1}), the absence of inclusions in the respective direction is captured by
\begin{equation}\label{eq.perco_matrix}
\begin{array}{rl}
  \WH p^{\rm M}_d &= 1-\min (\ull b_d \ast \ull{\rm{RVE}})\\
  &= \max\big( \ull b_d \ast (1-\ull{\rm RVE} )\big) \,,
\end{array}
\end{equation}
where $1-\ull{\rm RVE}$ is a phase inversion of the image in the bi-phasic setting.
Generally, ${(p^{\rm I}_d -1) \neq p^{\rm M}_d}$, due to the minimum/maximum operation.

The band feature detectors introduce hyperparameters, one being the directions to be considered with the band feature detectors (see \figref{fig.band_features} a)\,), and one being the width of the band feature detector, which controls the \textit{minimum connection width}. In this study, we chose the width of the band feature detectors to be 4px, and the angle increments of the band feature detectors are set to be $\frac\pi8$ for both phases.

\textit{Remark:} An extension of the approach to multiphase materials can be gained by using the band features on the individual phase indicator functions.

The novel band feature coefficients have been used to enrich the existing feature set, and a new model was trained. After convergence, the model is once again used to predict the test set and samples of high aleatoric uncertainty are displayed in \figref{fig.special_samples_2}. It can be seen that the samples leading to high aleatoric uncertainty now often contain dispersed inclusions which are spread over the whole RVE. It can also be seen that the found inclusions are comparatively small with some overlap. If multiple particles are connected this often leads to non-convex inclusion clusters. Consequently, we seek features that are able to characterize inclusion dispersal and inclusion shape.

\begin{figure}[h]
    \centering\includegraphics[width=0.48\textwidth]{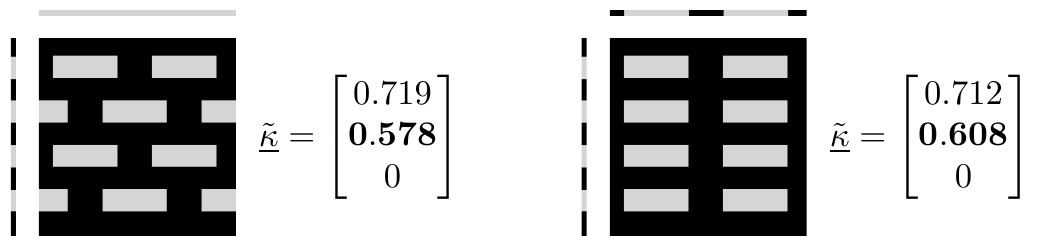}
  \caption{Two artificial microstructures are shown with their respective effective heat conductivity to the right. On the top/left the respective reduced line projections are graphically given. The RVEs only vary in the horizontal position of half the inclusions but differ noticeably in their effective response.}\label{fig.projected_wall}
\end{figure}

%% Full page figures moved to this part such that they are on the page of the correct chapter 
\begin{figure*}%[tp]
    \centering\includegraphics[width=\textwidth]{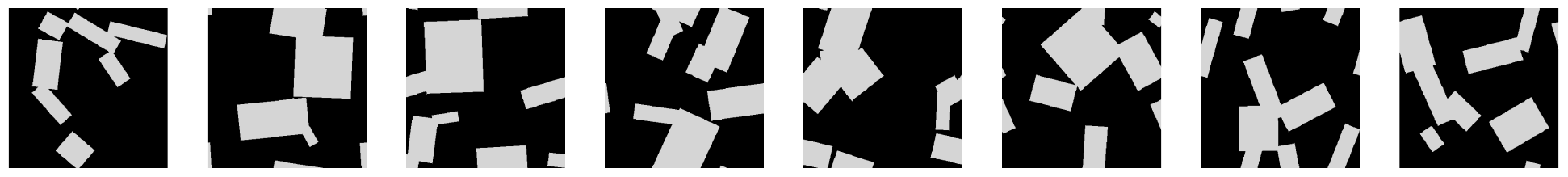}
  \caption{Iteration 3: A selection of samples which led to the high relative prediction error and high aleatoric uncertainties are shown. These samples were found by a model trained using all the previously mentioned features.}\label{fig.special_samples_3}
\end{figure*}

Physically motivated the next feature descriptor aims to quantify global flux hindrance, i.e., if there are multiple inclusions which form a disconnected barrier in a certain direction. 
This measure is computed by reducing the 2d-image to a 1d-line and noting if there is at least one voxel containing the inclusion phase. An example is illustrated in \figref{fig.projected_wall}, where the full line (on the top left) indicates the \textit{disconnected barrier}. Taking the average of the line, we obtain a scalar valued feature for each direction. This is implemented in x- and y-direction via (given in \texttt{python} pseudocode)

\begin{equation}\label{eq.projected_wall}
  w_i = \text{mean\Big(  sum( $\ull{\rm RVE}$, axis$=$}i) \geq 1  \Big)
\end{equation} 
where $\geq 1$ checks if there is at least 1 pixel found and the sum operator is conducted in axis/dimension $i$, following the syntax of \texttt{numpy.sum}.

\begin{figure}[h]
  \centering \includegraphics[width=0.48\textwidth]{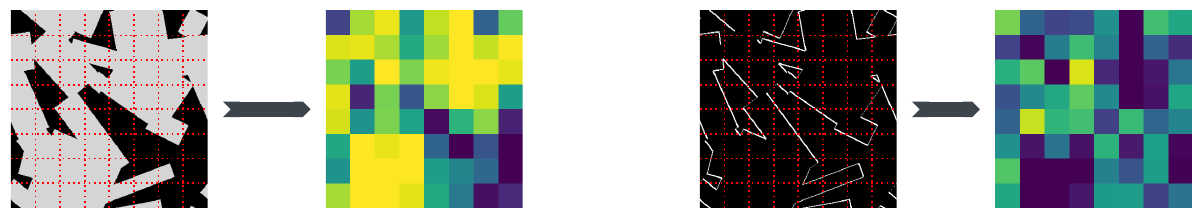}
  \caption{The process of extracting the distribution of volume fraction/edges is shown above, reducing the $400\times 400$ image data to a $8\times8$ distribution with the chosen parameters. On the left the volume fraction distribution is shown, and on the right the edge distribution for $k^2$ is shown.}\label{fig.vol_distribution}
\end{figure}

To quantify inclusion dispersal and approximate size of inclusions the RVE was subdivided into a grid of local regions/cells. The local volume fraction~$c_{IJ}$ ($I, J$ in the coarse grid) was computed, coinciding with \textit{average pooling} \citep{gholamalinezhad2020pooling} (cf. \secref{sec.conv_net}). The local volume fraction on the coarse grid yields a spatial distribution of the relative amount of inclusion phase, see \figref{fig.vol_distribution} left.

Taking the mean of $\ull{c}$  the global volume fraction is obtained, i.e., average pooling is volume preserving. New insights can be gained by taking the standard deviation~$\sigma^f (\ull c)$ and skewness~$\WT{\mu}^f_3(\ull c)$ of the distribution, which approximately represents the size of inclusion clusters. Additionally, the number of cells containing a certain volume fraction were counted. Firstly, cells containing only one phase were counted, i.e., cells with the local volume fraction of 0 or 1. The remaining cells were further subdivided into thirds, such that the cells were counted as
\begin{equation}\label{eq.local_volume}
  \ul c = \frac1{n_{\rm cells}} \left\lbrace   
  \begin{array}{lcrccr}
    \# \big( &&f_{\rm cell} &<& \epsilon &\big)\\[2pt]
    \# \big( &\epsilon &\leq  f_{\rm cell}&<& \frac13 - \epsilon&\big)\\[2pt]
    \# \big( &\frac13 - \epsilon &\leq  f_{\rm cell} &<& \frac23 - \epsilon&\big)\\[2pt]
    \# \big( &\frac23  - \epsilon &\leq  f_{\rm cell} &<& 1 -\epsilon &\big)\\[2pt]
    \# \big( & 1 - \epsilon &\leq  f_{\rm cell}& & &\big)
  \end{array}\right.,
\end{equation}

where $\#(\bullet)$ denotes the operation counting the number of cells fulfilling the condition, $f_{\rm cell}$ being the local cell volume fraction ($c_{IJ}$) and $\epsilon \leq \frac1{n_x\cdot n_y}$ a numerical tolerance. 
Here one more hyperparameter is introduced, namely the window size of the average pooling (we chose $\frac18n_x\times\frac18n_y$ in this study), which determines the size of the coarse grain voxels. In addition to that one could also consider a different partitioning for the counting operation where we fixed it to be divided into equal thirds.

\begin{table*}%[tb]
  \caption{The proposed features, are summarized in the table. The described phenomena of each feature as well as number of features and hyperparameters is shown. Note that when changing some hyperparameters the number of features may also change.}\label{tab.new_features}
\centering{\footnotesize
\begin{tabular}{|l|c | c | l |c| } 
\hline 
 \begin{minipage}{2.9cm}feature name \end{minipage} 
&\begin{minipage}{1.5cm}\centering  number of features \end{minipage} 
&\begin{minipage}{2.2cm}\centering  number of hyperparameters \end{minipage} 
&\begin{minipage}{3.5cm}captured phenomena \end{minipage} 
&\begin{minipage}{2.9cm}\centering  related equation \end{minipage} \\[2mm]
  \hline
 & &&&\\[-2mm]

 band features & 16 & 2 & linear phase connectivity & \eqref{eq.perco_incl}, \eqref{eq.perco_matrix}\\
 global directional mean & 2 & 0 & global flux hindrance & \eqref{eq.projected_wall}\\
 volume fraction distribution & 7 & 2 & inclusion size and dispersal & \eqref{eq.local_volume}\\
 directional edge distribution & 12 & 1 & inclusion shape and size &\eqref{eq.edges},\eqref{eq.edge_kernels}\\
\hline 
\end{tabular} 
} \end{table*}

Another feature descriptor is proposed to further quantify the size and sphericity of the inclusions through the surface information, since the ratio of $\frac{\rm inclusion\, area}{\rm inclusion \,perimeter}$ assist in the estimation. The ratio grows with inclusion size and is generally larger for circles than rectangles.
The surface information can be found via a convolution, where edge detectors like the Sobel operator \citep{vernon1991machine} are able to detect the (directional) surfaces of inclusions. The feature map of edges $\ull E^e$ can be computed as 
\begin{equation}\label{eq.edges}
  \ull E^e = \lvert{\ull k^e} \ast \ull{\rm RVE}\lvert, \qquad e\in\{1,2,3,4\}
\end{equation}
for different edge detectors $\ull k^e$. The chosen edge detectors are the horizontal, vertical and a diagonal Sobel filters, i.e.,
{\small
\begin{equation}\label{eq.edge_kernels}
  \begin{array}{rl@{\qquad}rl}
    \ull k^1 &= \begin{bmatrix} -1& 0& 1 \end{bmatrix}\,, & %\hspace{1.0cm} &
      \ull k^2 &= \begin{bmatrix} 0& -0.5 & -1\\ 0.5& 0 & -0.5 \\ 1& 0.5& 0 \end{bmatrix}\,, \\[6mm]%\hspace{1.0cm}
      \ull k^3 &= \begin{bmatrix} -1& -0.5 & 0\\ -0.5& 0 & 0.5 \\ 0& 0.5& 1 \end{bmatrix}\,, &%\hspace{1.0cm} 
    \ull k^4 &= \begin{bmatrix} -1\\ 0\\ 1 \end{bmatrix}\,. 
  \end{array}
\end{equation} }
The absolute value is taken in \eqref{eq.edges} since the orientation of the surface normal is irrelevant for our investigations.
Similar to the local volume fraction the edge feature maps are processed with average pooling, leading to a coarse grid representation~{${ \ull S^e}$}. The mean~$\mu^E(\ull S^e)$, standard deviation~$\sigma^E(\ull S^e)$ and skewness $\WT\mu_3^E (\ull S^e)$ is taken, leading to three additional features per edge detector. The same window size as for the local volume fraction $\ull{c}$ was used, which constitutes one hyperparameter for this feature. Additionally, other edge detectors, e.g., the Laplacian, could be considered.\\

With the auxiliary features at hand a new BNN model was trained for the next iteration, with all of the previously used and newly introduced features. Once again the RVE samples of high aleatoric uncertainty and high prediction errors ares shown in \figref{fig.special_samples_3}. As can be seen in most of the samples, clusters of inclusions are formed with narrow connections, leading often to non-convex edges. Similarly, inclusions do almost form a cluster with a few pixels distance in between them. No efficient method to quantify these phenomena was found, and other features which have been tested did not notably improve the models prediction. Thus, the feature engineering was halted at the third iteration of feature engineering.

In total $37$ new features have been proposed by utilizing four different feature descriptors, which are compactly summarized in \tabref{tab.new_features}. Each of the features is either obtainable via convolution, or obtainable by computationally efficient operations.
\subsection{Convolutional neural networks}\label{sec.conv_net}

Convolutional neural networks (Conv Nets) \citep{o2015introduction} are derived from feed forward neural networks \citep{basheer2000artificial,lissner2019data} to make the classification/regression of high-resolution images tractable via weight sharing \citep{lecun1989generalization}. The evaluation of Conv Nets is similar to dense feed forward neural networks and conducted in one forward pass. The matrix multiplication used in dense feed forward neural networks is replaced by a convolution
\begin{equation}
 \ull A_{l+1} = f_l( \ull W_l \ast \ull A_l + b_l )\,,
\end{equation}
with $\ull A_{l+1}$ being the image output of layer $l$, $\ull W_l$ denoting the kernel, $b_l$ the element wise added bias/offset, and $f_l$ the activation function. Here the kernel $\ull W_l$ and the bias $b_l$ are the trainable parameters of the neural network.

\begin{figure}[h]
  \centering \includegraphics[width=0.45\textwidth]{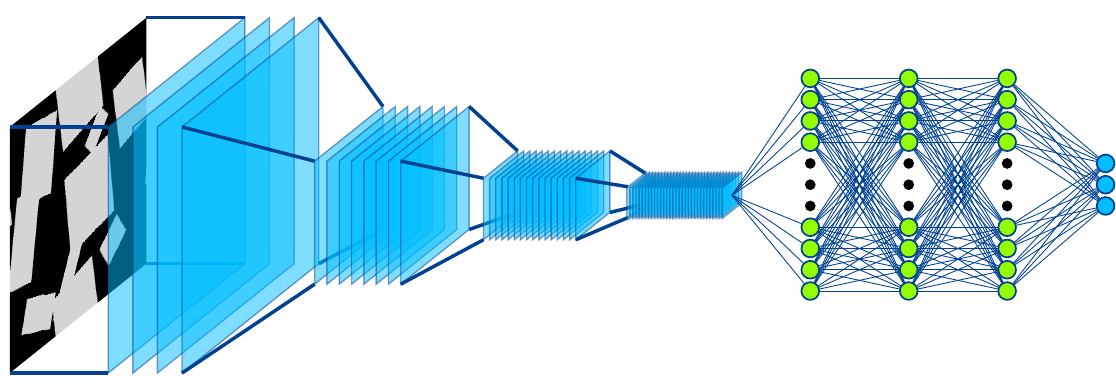}
  \caption{A schematic overview of a vanilla convolutional neural networks is shown. The supposed feature extraction is conducted in the convolutional layers (displayed as the light blue rectangles), where each rectangle symbolizes one channel. The last convolutional layer is followed by a flattening operation and a subsequent dense feed forward neural network (displayed with the green neurons).}\label{fig.conv_net}
\end{figure}

One major advantage of Conv Nets over dense neural networks is that the small kernel (e.g. $3\times3$ is a popular choice) is used to scan over the entire image, processing the entire information with only few parameters.
To further improve information processing, multiple \textit{channels} per layer can be specified (illustrated through multiple slices in \figref{fig.conv_net}), where each channel represents a different feature map (of same resolution). This leads to kernel matrices being of size ${m_y\times m_x\times n_{\rm in}\times n_{\rm out}}$, having different kernel weights for each input-output channel combination. Thus, each output channel of the current layer is obtained by processing all input channels with independent kernels, which results are generally averaged.

\begin{figure}[h]
\centering \includegraphics[width=0.40\textwidth]{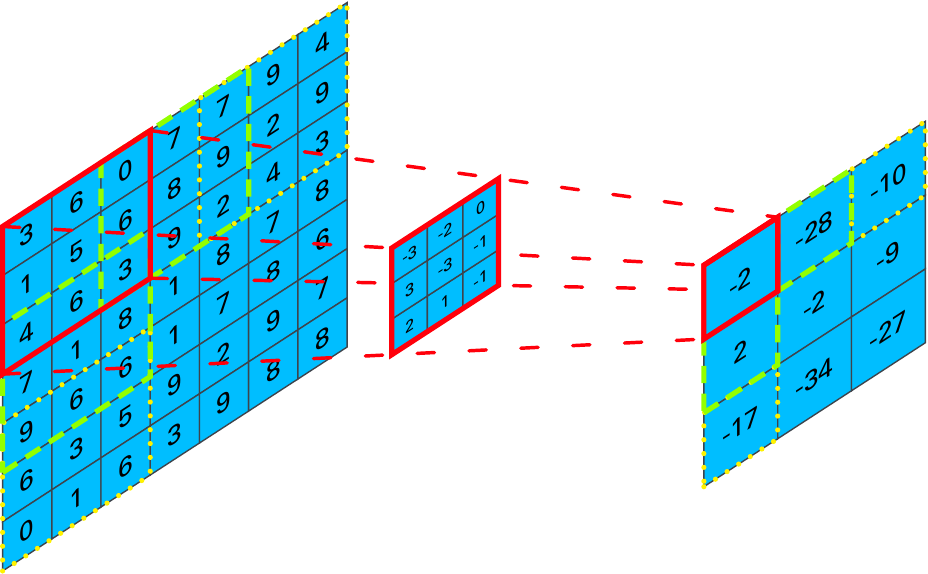}
  \caption{A convolution using a $3\times 3$ kernel (displayed in the middle) with stride 2 applied to a $7\times7$ image (left) is shown. The resulting image of size $3\times3$ after convolution is given to the right. Each colored box denotes the discrete positions of the kernel for the convolution operation, matched on input and output image.}\label{fig.convolution}
\end{figure}

To keep the computational overhead feasible, downscaling of the image resolution is conducted as more channels are added, which is generally achieved via \textit{stride} or \textit{pooling}. The stride is implemented through the convolution, denoting the stepping width of kernel evaluation interval, i.e., if stride=2 then the kernel gets evaluated on every other position, and a downscaling of approximately factor two is achieved along each dimension (\figref{fig.convolution}).

\begin{figure}[h]
  \centering \includegraphics[width=0.40\textwidth]{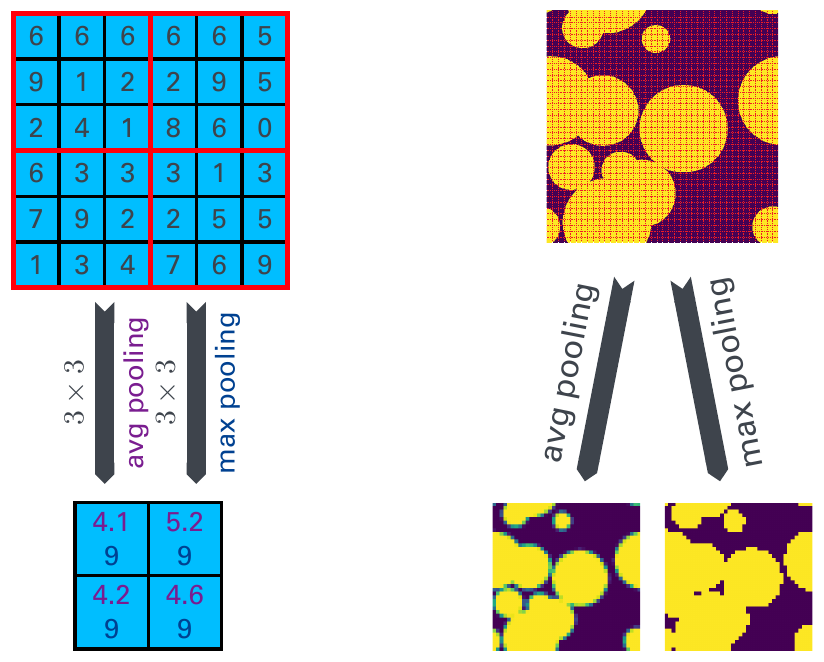}
  \caption{Average and max pooling is exemplarily displayed for an arbitrary $9\times9$ image on the left, where the discrete values after the operation are shown in the respective colors. The red rectangles highlight the spatial field considered for each output value. On the right ${10\times10}$ average and max pooling was applied for an exemplary microstructure image. The pooling 'grid' is shown on the image in red, where each grid corresponds to one output pixel.}\label{fig.pooling}
\end{figure}

A \textit{pooling} layer replaces the convolution operation with e.g. the maximum or average operation (\figref{fig.pooling}), achieving downscaling analogously through the stride. In a general case the stride in pooling is set equal to the size of the pooling \textit{kernel}. Interpreting the different pooling operations one can see that max pooling singles out the peak activation of the kernel in the local neighbourhood, whereas average pooling reflects the average occurrence of the feature in a local region. After a pooling layer, the number of channels is kept constant meaning that each channel is pooled individually. 

Another special operation in Conv Nets are ${1\times1}$ convolutions, which are used to control/reduce the number of channels while generally retaining the spatial resolution (with a stride$=1$). Additionally, a nonlinear activation function can be deployed after the ${1\times1}$ convolution.\\

With these operations at hand the convolutional layers can be built, where the schematic overview of a Conv Net in \figref{fig.conv_net} has four convolutional layers, where each layer except the first one has a stride$\gt$1 and an increasing number of output channels per layer.
After the last convolutional layer, the derived features in the \textit{spatial image representation} are flattened into a 1d-vector and a dense feed forward neural network is deployed to conduct regression/classification. Hence, the convolutional channels are often interpreted as feature extractors, whereas the dense neural network is utilized for the prediction based on the features extracted by the convolutional layers. Such neural networks will be referred to as \textit{generic} Conv Nets below.

\subsubsection*{Padding in Conv Nets}
The convolutional kernels derive features based on neighbourhood information and consequently, on the boundary of the image where the neighbourhood is undefined, a loss of information is introduced. To enable the Conv Net to correctly consider the full neighbourhood relationship, periodic \textit{padding} before each convolution operation was deployed. This has been implemented in \texttt{tensorflow} similar to \citep{schubert2019circular} and is made publicly available in \citep{codebase}, where also the implementation for the data augmentation scheme is found.

\begin{figure}[h]\centering
  \includegraphics[width=0.18\textwidth]{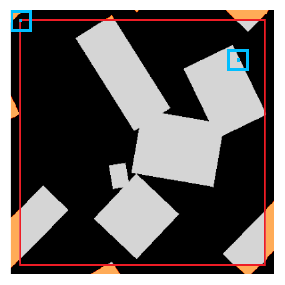}
  \caption{The original RVE (inside red box) is padded by half the kernel size. The padding is highlighted in orange. The blue rectangles show the considered region while evaluating the convolution with a kernel at the dot position for two exemplary convolution locations.}\label{fig.padding}
\end{figure}

For illustration a periodically padded image is shown in \figref{fig.padding}, where it can be seen that the padding ensures the spatial resolution to stay constant after the convolution operation, since the kernel evaluation is only defined within the red box of original spatial resolution.
Note that in a deep Conv Net periodic padding is deployed in each layer.

\subsubsection*{Data augmentation}

\begin{figure}[h]
  \centering
    \includegraphics[width=0.35\textwidth]{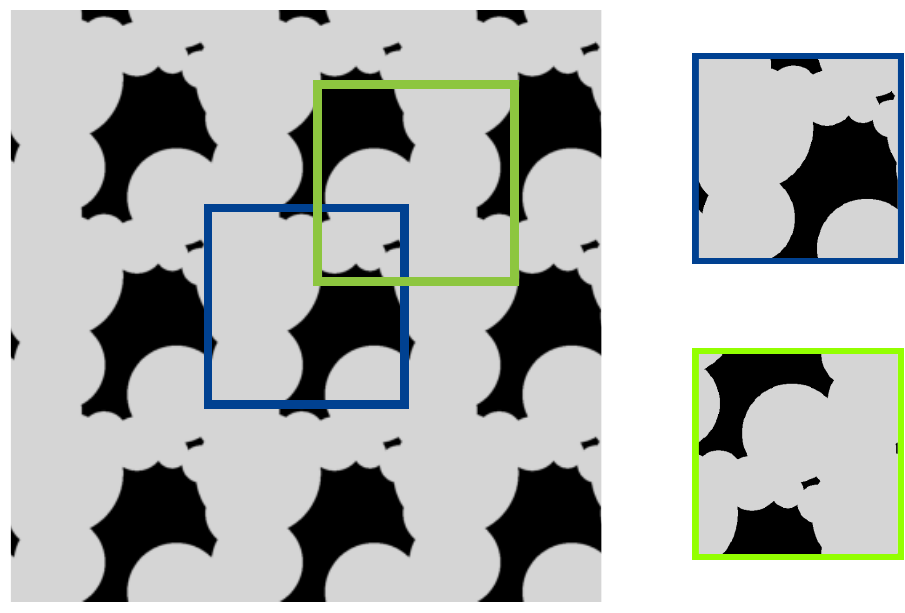}
  \caption{The blue box highlights the originally generated RVE which characterizes the macroscopic material by a repeated continuation in each direction. The green box outlines a completely different image of the same RVE obtained from translation in the periodic medium.}\label{fig.rve_translation}
\end{figure}

Data augmentation aims at increasing the size of the training set, which inherently has a regularizing effect \cite{kauderer2017quantifying}. 
One study has successfully applied cutout, which goes as far as to simply gray out larger regions of the input image during training \citep{devries2017improved}. We aim to virtually increase the size of the training set by translating the frame of the RVE, which does not alter the macroscopic material due to assertion of periodicity (\figref{fig.rve_translation}). This additionally assists the Conv Net in learning the translational invariance of the RVE frames, and can generate up to $400^2$ (i.e. image resolution) snapshots/samples of the input image per RVE. For memory considerations, this is implemented 'online' during training, where 50\% of the RVE frames are randomly translated every 10-th epoch.

\subsection{Deep inception module}\label{sec.inception}

In the context of microstructure homogenization, the material behaviour is impacted by various factors, e.g., locations with narrow gaps between inclusions, as well as large inclusion clusters. In the special case of percolation, it is relevant that one inclusion cluster stretches over the entire RVE, often in a curved manner, which also has to be detected in order to make accurate predictions of the materials behaviour. Thus, different characteristics with different relative size in the image have to be detected by the convolutional layers.%, which implies that differently sized kernels, operating on the same input image, will assist the Conv Net in conducting accurate predictions.

\begin{figure}[h]
  \centering\includegraphics[width=0.36\textwidth]{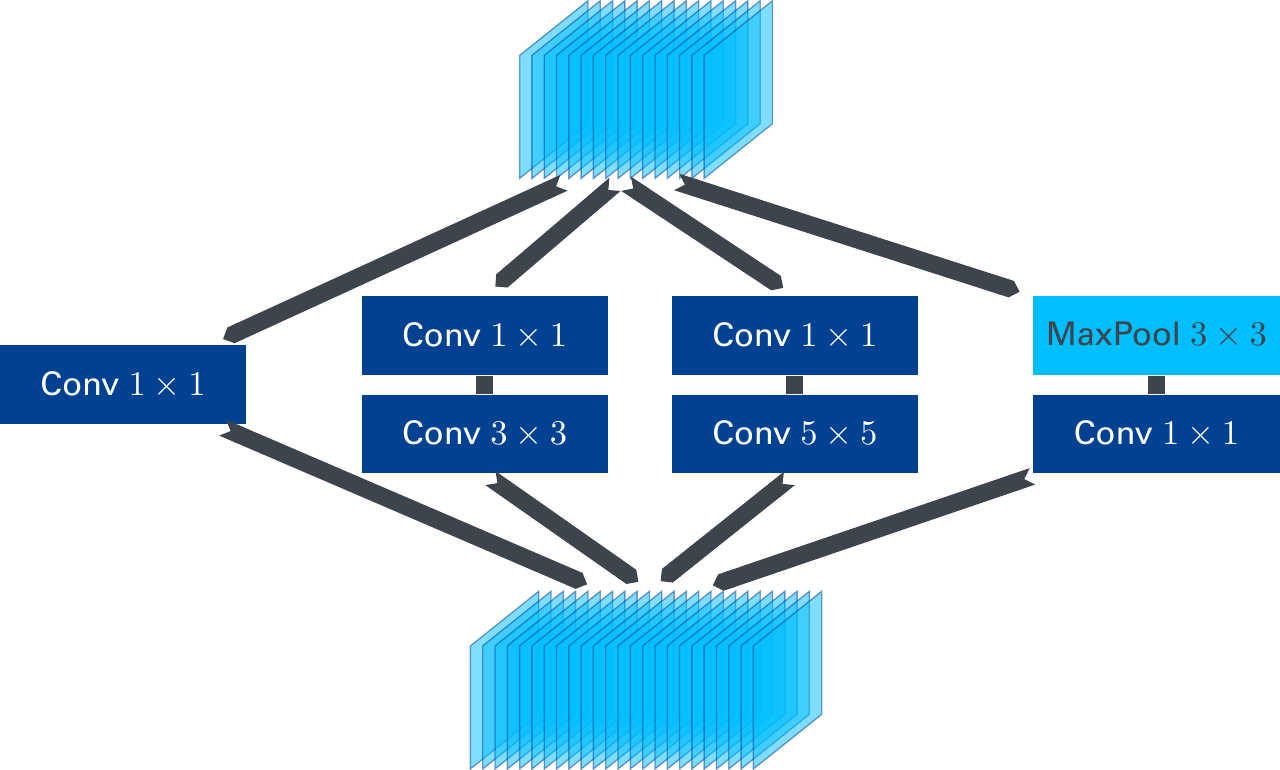}
  \caption{The original inception module is graphically illustrated where multiple convolution operations are conducted between the layers. The graphic layout is copied from \citep{szegedy2015going} fig.2 and slightly modified.}\label{fig.inception_orig}
\end{figure}

In order to detect these various features, we deploy parallel branches of differently sized kernels. Thereby, we design the Conv Net to capture differently sized characteristics. Something similar, namely the inception module has been previously implemented by \cite{szegedy2015going}. Their original intention was to reduce computational overhead and memory usage by increasing width instead of depth in Conv Nets. Their resulting model, which only used inception modules (\figref{fig.inception_orig}), achieved state of the art results.

The inception modules were implemented by replacing the convolution operation between two layers by multiple parallel convolutions of differently sized kernels, where each of the convolution operations has padding and a stride of 1, since the feature maps are concatenated channel wise at the end of the module. To reduce the spatial resolution in the deep Conv Net, \cite{szegedy2015going} used pooling between multiple inception modules.

\begin{figure}[h]
  \centering
  \includegraphics[width=0.45\textwidth]{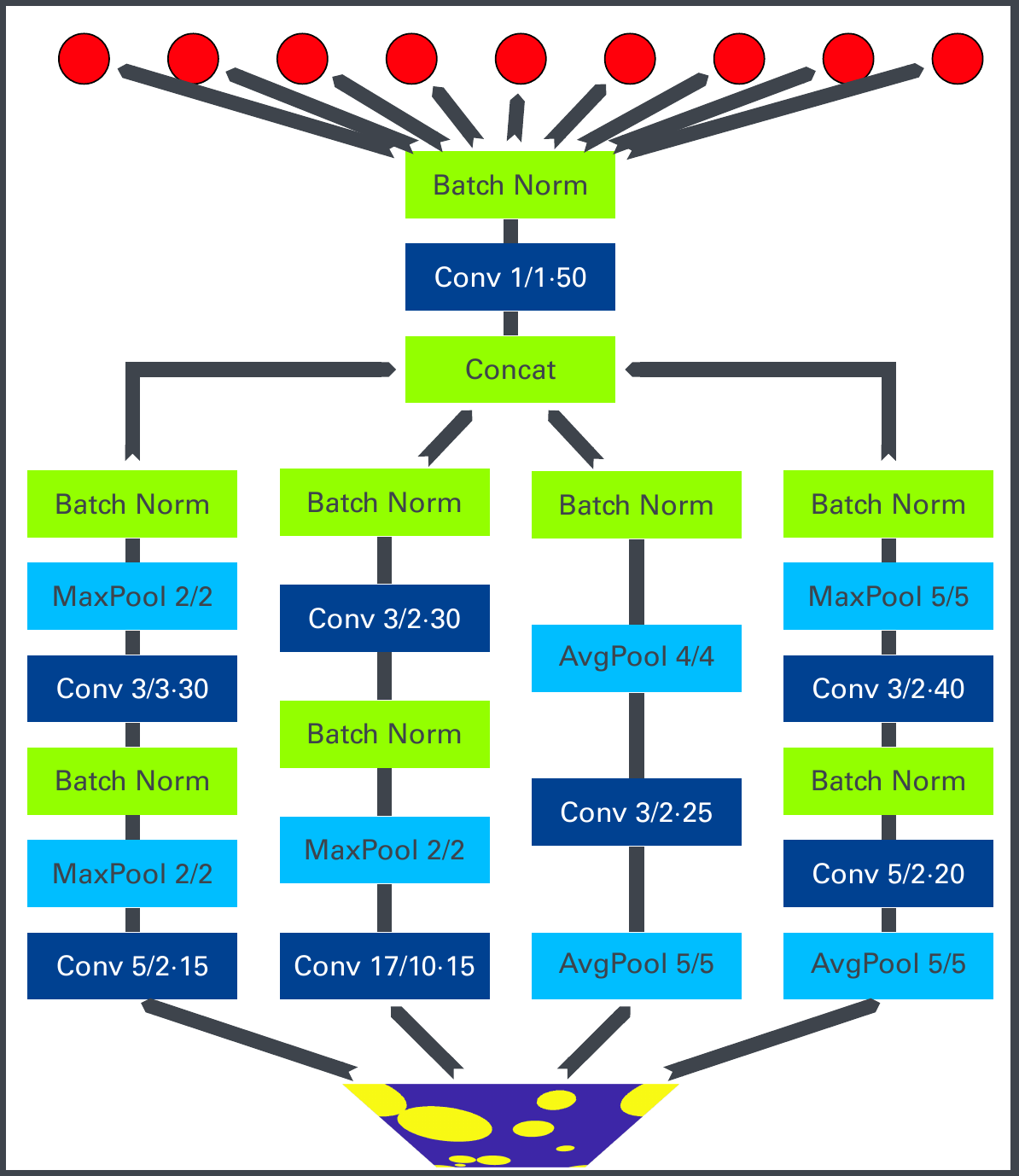}
  \caption{One deep inception module is shown where multiple deeper convolution branches are conducted in parallel. Each kernel is quadratic and the dimensions are read as size/stride $\cdot n_{\rm channels}$.}\label{fig.deep_inception_module}
\end{figure}

Building upon the idea of multiple parallel convolutions, we increase the depth in each parallel convolution \textit{branch}, and propose the \textit{deep inception modules} (\figref{fig.deep_inception_module}). The constraint of stride$=1$ in the convolutional layers is dropped, and the number of operations in each branch can also be flexibly adjusted. Then, we design each branch to capture different sized phenomena, e.g., by a preceding average pooling/coarse graining\footnote{Note that large average pooling introduces only a minor loss in information, since edge information is mostly retained when using float values (c.f. \figref{fig.pooling})}, we can easily increase the receptive field of the first $5\times5$ convolution to be $25\times25$ pixels (\figref{fig.deep_inception_module} rightmost branch). Additionally, we ensure that there exist also branches which deploy convolution operations directly using the raw image information, to capture small sized effects within the RVE (\figref{fig.deep_inception_module} leftmost branch).

The only remaining constraint in the deep inception modules is that the downsampling factor through stride/pooling has to match in each branch if the channels are concatenated after the convolution operation.  In \figref{fig.deep_inception_module}, each branch individually achieves a downsampling of a factor of $40$, and the channels of each branch are concatenated at the end.
\subsection{The hybrid neural network}\label{sec.hybrid_model}

\begin{figure*}%[!hp]
  \centering
  \includegraphics[width=0.76\textwidth]{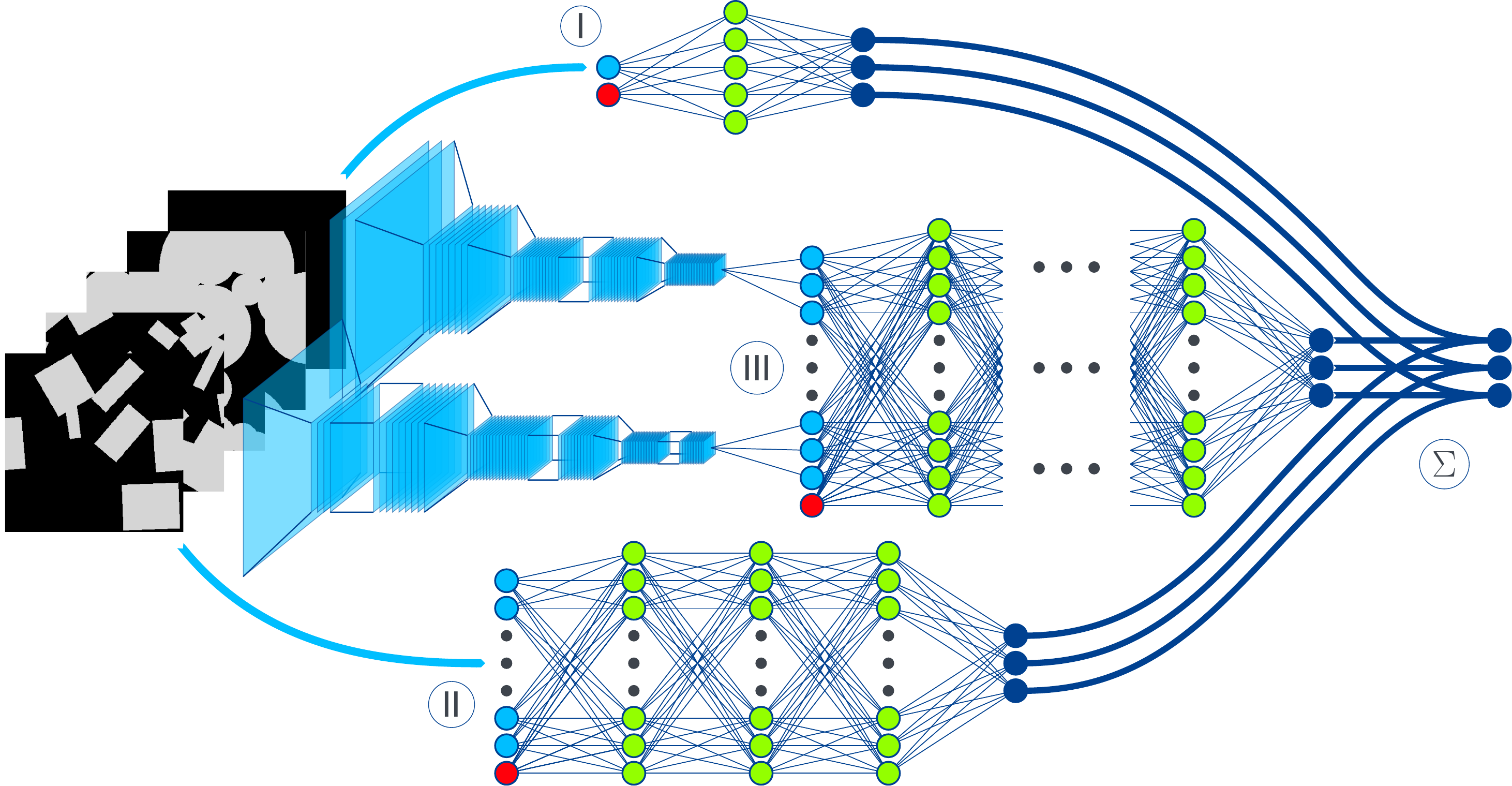}
  \caption{The developed \textit{hybrid neural network} utilizes parts of a generic dense feed forward neural networks as well as parts of a Conv net with deep inception modules. The prediction of the effective property is obtained by summing up the prediction of each sub-part of the model.}\label{fig.hybrid_model}
\end{figure*}

\begin{figure}[h]
\centering\includegraphics[width=0.5\textwidth]{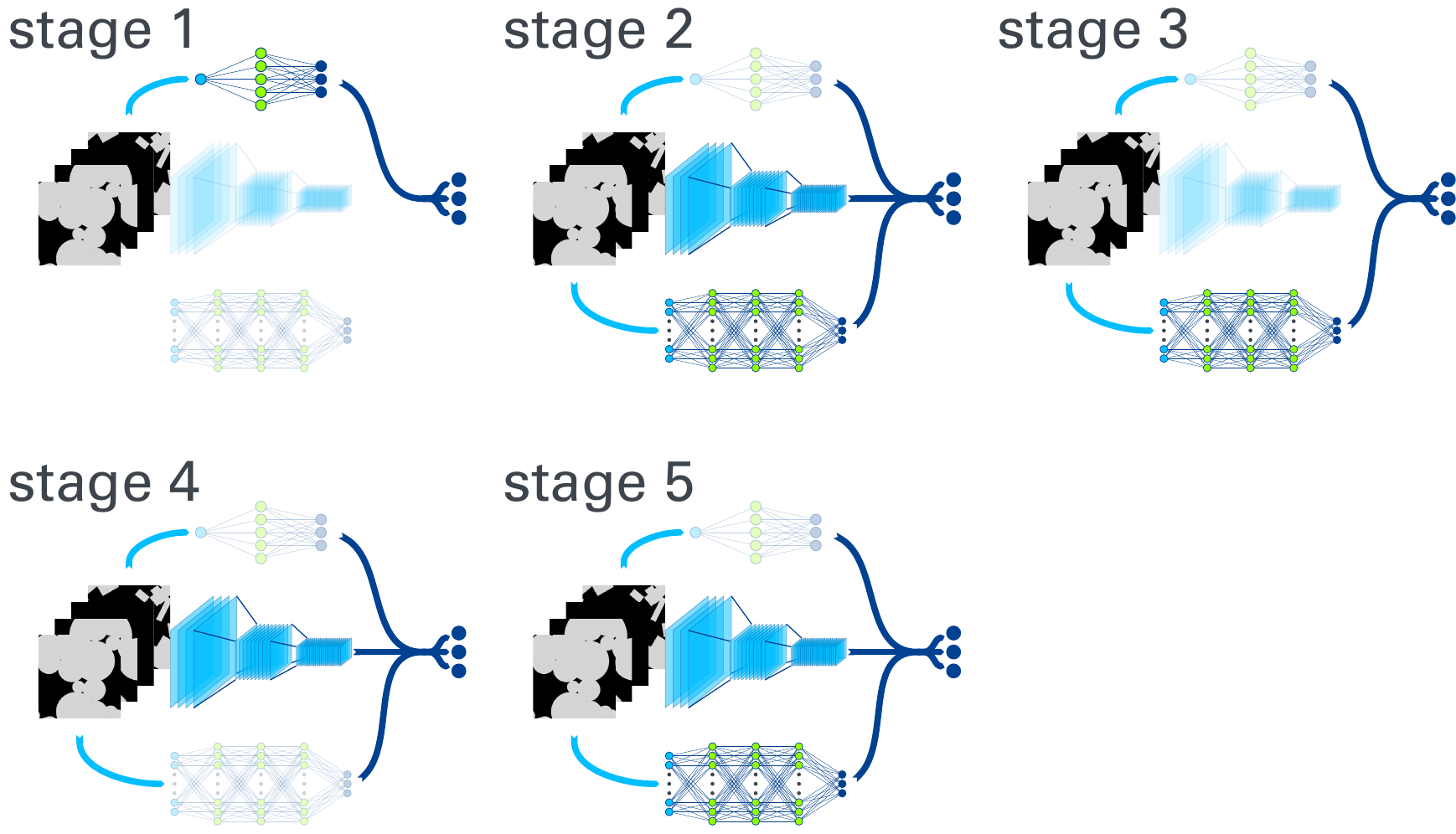}
\caption{The multiple stages of the hybrid models training are shown. If a model subpart is transparent in a stage, then gradients were not computed thereof. The line connections represent which part of the model contributed to the prediction in the current stage. Each stage except for stage 2 is trained to convergence.}\label{fig.multistage_training}
\end{figure}

So far two different model archetypes, i.e., the dense feed forward neural network (FFNN) using the handcrafted features and the convolutional neural network (Conv Net) are introduced to predict the homogenized material property. Both model archetypes are able to outperform the reference model \cite{lissner2019data} and we aim to obtain further improvements by combining the FFNN and Conv Net in parallel. 
To additionally support the prediction, a bypass using only the volume fraction, which is the single most impactful variable in homogenization, has been implemented to serve as a baseline prediction. The resulting hybrid neural network is shown in \figref{fig.hybrid_model}, which consists of three different contributions to the models prediction. The models prediction is given as

\begin{equation}\label{eq.full_prediction}
  \tilde{\ul\kappa} = \tilde{\ul\kappa}_{\rm vol} + \Delta( \tilde{\ul\kappa}_{\rm features} + \tilde{\ul\kappa}_{\rm Conv Net})\,, 
\end{equation}
where each predicted subpart $\hat{\ul\kappa}_\bullet$ denotes one subbranch of the hybrid neural network. The idea is that each branch considers increasingly high level features in order to predict the effective heat conductivity, which is governed by complex geometrical effects. To assist the models convergence, a multistage training is implemented. The entire scheme is graphically illustrated in \figref{fig.multistage_training}. 

Firstly, branch (I) using the volume fraction is trained independently until convergence (stage 1). After convergence the parameters of (I) are frozen, but the branch will contribute to all subsequent predictions.
As a next step, the entire model is trained for only a few epochs (stage 2), to move the model in the proximity of a local minimum. 
Since the handcrafted features (\secref{sec.new_features}) remain static during training, branch (II) using these features is trained until convergence (Stage 3), which trainable parameters are frozen thereafter. In the next step the last branch (III) containing the Conv Net is trained until convergence (stage 4), while the feature branch (II) contributes to the prediction. The motivation is that the Conv Net can capture high level features which explain the remaining variations uncaptured by the handcrafted features. In the final step, the interdependent parameters are finetuned by once again training the entire model until convergence (except for branch (I)).

\begin{figure*}%[bp!]
  \centering
  \includegraphics[width=\textwidth]{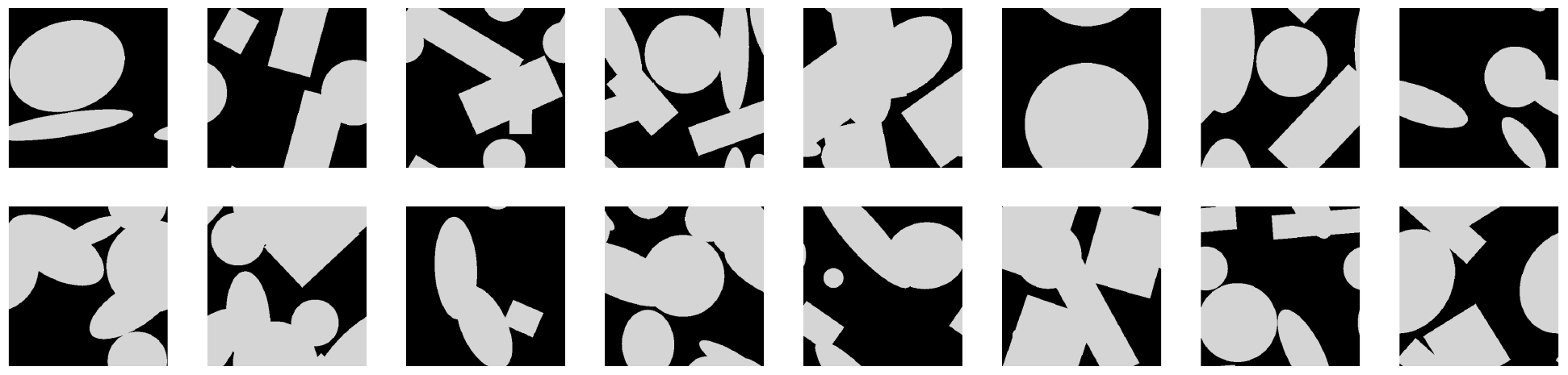}
  \caption{A few samples out of the \textit{benchmark} set are displayed. Note that the combination of different inclusions within one RVE is unseen to the machine learning algorithm and represents truly unseen data. Some samples of the training data were shown in \figref{fig.data_overview}. }\label{fig.benchmark_rve}
\end{figure*}

\section{Results}\label{sec.results}

During the process of finding the best model, i.e., the hybrid neural network, a multitude of neural networks has been trained in the step by step procedure. The results of each intermediate step are compactly summarized in \tabref{tab.model_errors} and visualized via a R$^2$ plot in the appendix \ref{fig.rsquare_plots}. Further details on the different prediction contributions, i.e., the handcrafted features and the convolutional neural networks (Conv Nets) are given separately.
In general, for the different network archetypes, different losses have been deployed. Regarding model parameter optimization, the same state of the art optimizers have been used across the different models. This consists of ADAM gradient back propagation with weight decay \cite{loshchilov2017decoupled}, early stopping and a constant learning rate with default hyperparameters. The model training has been implemented using the \texttt{tensorflow} API \cite{tensorflow2015whitepaper}, and the code for training and post processing can be found in \cite{codebase}. 

All models have been trained using the same data outlined in \secref{sec.data} (elaborated in appendix \ref{appendix.data}). Validation was done using a benchmark dataset of completely unseen inclusion shapes whilst a mix of inclusions within a single RVE is allowed (\figref{fig.benchmark_rve}). As an intuitively interpretable metric we use the relative error measure: 
\begin{equation}\label{eq.rel_mse}
  e = 100\cdot \sqrt{\frac1{n_{\rm samples} \cdot n_\kappa} \sum\limits_{i=1}^{n_{\rm samples}} \sum\limits_{j=1}^{n_{\kappa}} \frac{ (y_{ij} -\hat y_{ij})^2}{y_{ij}^2}}\,,
\end{equation}
which coincides with the error measure introduced in \eqref{eq.rel_norm} and will be denoted with \relmse in the future. Here $y_{ij}$ denotes the target values, $\hat y_{ij}$ denotes the models prediction for all samples $i$ and all components $j$ (of the heat conduction tensor).

\subsection{Feature engineering and selection}\label{sec.feature_selection}

\begin{figure*}%[tp]
   \centering\includegraphics[width=0.99\textwidth]{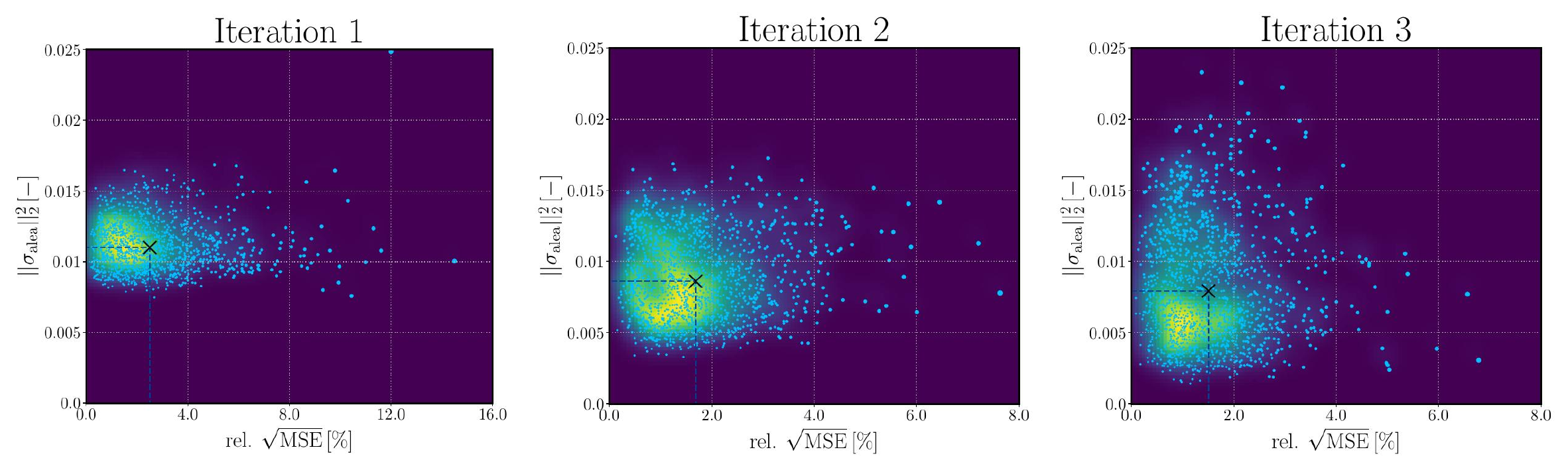}
  \caption{A density plot is shown which relates the aleatoric uncertainty to the \relmse in each iteration step, overlayed with a scatterplot of the actual predictions. Samples further away from the brightest spot in the density plot are displayed larger for better visibility. The 'x' marks the mean values. Note that the abscissa of the left most plot is differently scaled.}\label{fig.alea_prediction}
\end{figure*}

During the feature engineering approach, in each iteration step a new Bayesian neural network (BNN) was trained using the same architecture and the Bayesian loss described in \eqref{eq.bnn_loss}. The architecture is given in the appendix \ref{tab.feature_architecture}. The features which have been added in each iteration step were motivated and described in \secref{sec.new_features}. The actual improvements per step are given in \tabref{tab.model_errors} and are graphically shown in \figref{fig.alea_prediction}. In \figref{fig.alea_prediction}, the density blur highlights the locations where most of the model predictions coincide, and it can be nicely seen that the center of it, i.e., the brightest spot, shifts closer to the origin in each iteration. It can also be seen that the mean prediction error improves whilst the average predicted uncertainty also decreases.
Note that in the first iteration the coefficients of the 2-point correlation function (2PCF) has been used \cite{lissner2019data}, in the second iteration the band features were added, and in the third iteration the remainder of the features was added, consisting of the volume fraction distribution, the directional edge distribution and the global directional mean (c.f \tabref{tab.new_features}).

\begin{figure}[h]
  \centering
  \includegraphics[width=0.50\textwidth]{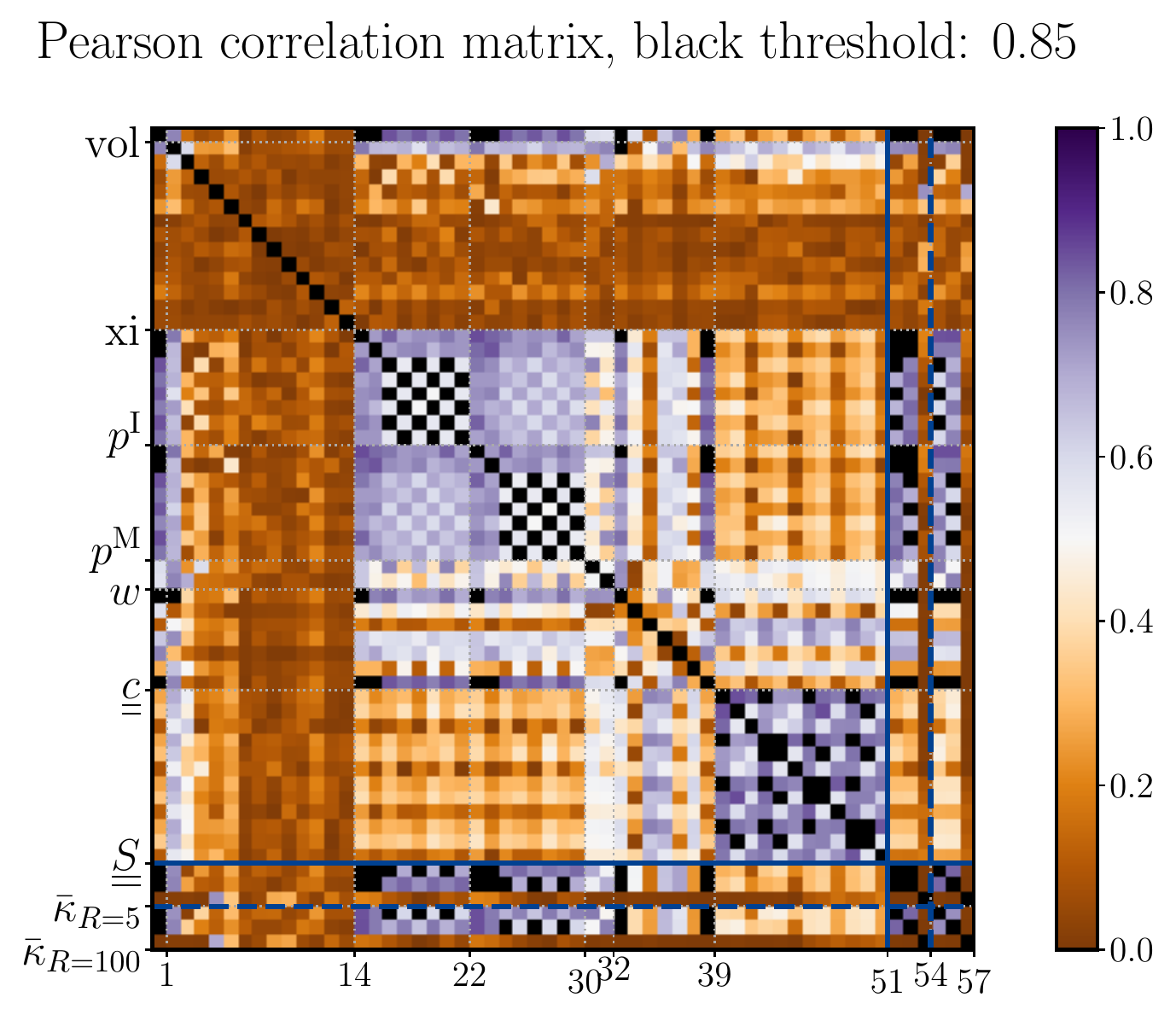}
  \caption{The absolute value of the Pearson correlation scores for all features as well as the effective thermal properties of 3000 samples are given above for the phase contrast of $R=5$ and $R=100$. Absolute values greater than 0.85 are blacked out to indicate high correlation. The marked labels on the ordinate note which feature is displayed up to the label (read from top to bottom, ordered as introduced in \secref{sec.new_features}), and the abscissa up to which index the current feature is given in the feature vector. The six features beyond the blue line denote the target values, i.e., the effective heat conductivity for the different phase contrasts.}\label{fig.correlation_features}
\end{figure}

As has been previously stated, any tests on additional features features yielded little to no improvements. Since the improvements from the second to the third iteration were smaller than in the first iteration, a feature selection process was initiated to spot if the introduced features are redundant or do not positively contribute to the prediction. The correlation matrix in \figref{fig.correlation_features} shows the absolute values of the Pearson correlation scores, indicates that the features are suitably uncorrelated, though some correlations remain within each feature class. 

In a further investigation step, the features have been ranked by two filter and wrapper methods. Filter methods estimate feature importance on plain data observation, and wrapper methods rate the importance based on an intermediate surrogate model which measures \textit{scores} of prediction contribution. The deployed filter methods used the Pearson correlation scores and the analysis of variance (ANOVA) via F-scores \citep{feir1974empirical}. The deployed wrapper methods were the recursive feature elimination (RFE)\citep{guyon2002gene} and a method deploying random forests \citep{geurts2006extremely}. Each of these methods found a slightly different order, presented as indices, given in arrays:

{\footnotesize{
\begin{equation}
  \begin{gathered}
    \text{\small{Pearson ranking:}}\\
    \begin{array}{lllllllllllllll} \large[ 
      &\cb{0},& \cb{38},& \clb{22},& \clb{14},& \clb{32},& \cb{24},& \cb{28},& \clb{16},& \cb{20},& \cb{26},& \clb{18},& \clb{15},& \clb{23},& \\
      &17,& 21,& 25,&  1,& 29,& 19,& 27,& 36,& 35,& 31,& 30,&  4,& 33,& \\
      &49,& 40,& 46,&  5,& 43,& 48,& 39,& 45,& 42,&  8,& 50,&  9,& \\
      &\co{37},& \co{3},& \cg{11},&  \co{2},& \cg{44},& \cb{47},& \cg{10},& \clb{12},& \cb{34},& \clb{41},&  \cb{7},&  \cg{6},& \cb{13} 
    \large] \end{array} \\[2mm]
    \text{\small{ANOVA ranking:}}\\
    \begin{array}{lllllllllllllll} \large[ 
      &\cb{0},& \cb{38},& \cb{22},& \clb{15},& \clb{14},& \clb{32},& \cb{24},& \clb{23},& \cb{28},& \clb{16},& \cb{26},& \cb{20},& \clb{18},&\\
      &29,& 21,& 17,& 25,& 1,& 19,& 27,& 36,& 33,& 35,&  4,& 30,& 37,&\\
      &31,&  2,& 48,&  5,& 39,& 42,& 45,& 46,& 8,& 40,&  3,& 11,&\\
      &\clb{49},& \cg{43},&  \cb{7},& \clb{41},&  \co{9},& \cb{13},& \cg{10},& \cb{34},&  \cg{6},& \clb{12},& \co{50},& \cg{44},& \cb{47}
    \large] \end{array} \\[2mm]
    \text{\small{forest ranking:}} \\
    \begin{array}{lllllllllllllll} \large[
       & \cb{0},& \cb{38},& \clb{23},& \clb{14},& \clb{22},& \clb{32},& \clb{15},& \cb{20},& \cb{24},& \co{25},& \cg{3},& \cb{26},& \cb{28},& \\
       & 18,& 17,& 16,& 21,& 29,& 27,&  2,& 19,& 33,& 31,& 30,& 37,& 4,& \\
       & 1,&  5,& 50,& 44,&  8,&  9,& 35,& 42,& 48,& 10,&  6,& 43,& \\
       & \cg{11},& \co{39},& \clb{41},&  \cb{7},& \cb{13},& \cg{45},& \cb{47},& \clb{12},& \cg{36},& \cg{46},& \co{40},& \clb{49},& \cb{34}
    \large] \end{array} \\[2mm]
    \text{\small{RFE ranking:}}\\
    \begin{array}{lllllllllllllll} \large[
        & \cb{0},& \clb{18},&  \cg{3},& \cb{26},& \cb{38},& \cb{20},& \cb{24},& \cb{28},&  \co{4},& \clb{16},&  \co{2},& \co{27},& \co{21},& \\
        & 14,& 22,& 19,& 15,& 23,& 17,& 39,& 25,&  1,&  8,& 31,& 29,& 50,& \\
        & 5,& 41,& 30,&  9,& 40,& 44,&  6,& 11,& 10,& 35,& 12,& 37,& \\
        & \cb{47},& \cg{43},& \cg{45},& \co{48},&  \cb{7},& \cg{46},& \co{32},& \clb{49},& \cb{13},& \co{33},& \co{42},& \cb{34},& \cg{36}
    \large] \end{array} 
  \end{gathered}
\end{equation} }}

where Python notation is used and the indices start at 0. Note that the first and the last 13 features are color coded to display agreement between the feature selection methods. Features marked by \cb{dark blue} have been ranked as the top/bottom 13 features by all methods, \clb{light blue} by three, \cg{green} by two, and \co{orange} by one.  
The utmost curious reader may compare the indices represented above with the indices and corresponding feature displayed in the correlation matrix (\figref{fig.correlation_features}). Overall, the volume fraction and most of the band features were rated as the most significant for all methods, and generally some edge distribution features and reduced coefficients of the 2PCF were commonly rated with the lowest scores.

\begin{figure}[h]
\centering  \includegraphics[height=0.20\textheight]{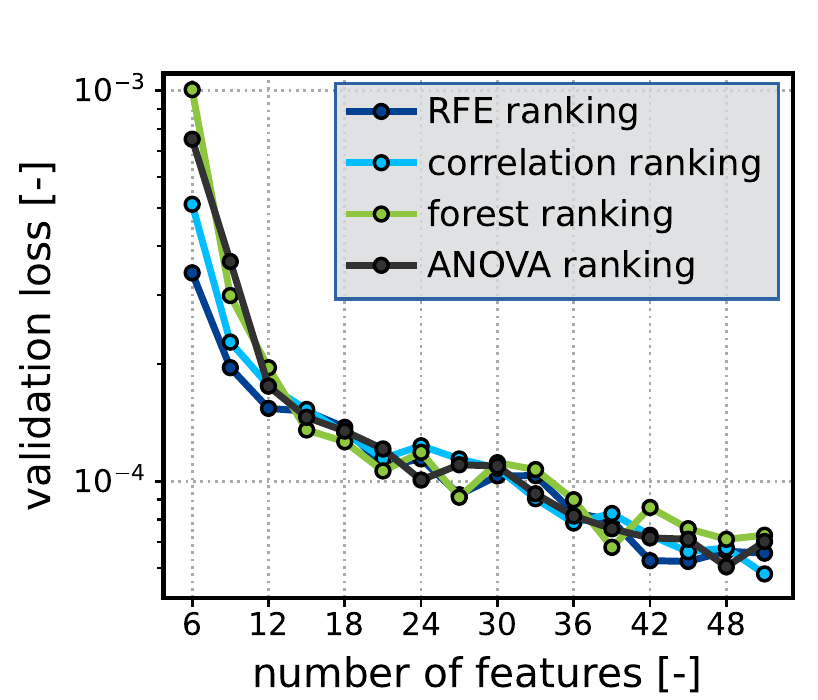}
  \caption{The validation loss development with respect to the total number of features is shown. Each scatter shows the lowest validation loss with the best BNN out of five for the different feature selection methods.}\label{fig.n_features}
\end{figure}

\begin{table*}%[tp] 
  \caption{Error measures are given for each step in the improvement process of the artificial neural networks. The full numbers present the error measures computed on the benchmark set and the values given in the brackets refer to the error measures computed on a test set affine to the training data. The different model types are separated via horizontal lines.}\label{tab.model_errors}
\centering{\footnotesize
\begin{tabular}{|l|cc cc rr|r| } 
  \hline 
 \begin{minipage}{2.6cm}Model type \end{minipage} 
 & \multicolumn{2}{c}{\begin{minipage}{3.0cm}\centering mean relative error [\%] \end{minipage}} 
 & \multicolumn{2}{c}{\begin{minipage}{3.0cm}\centering median relative error [\%] \end{minipage}} 
 & \begin{minipage}{1.15cm}\centering rel. \\$\sqrt{\text{MSE}}$ [\%] \end{minipage}
 & \begin{minipage}{1.15cm}\centering MSE \\ $10^{-5}$ [-] \end{minipage}
 & \begin{minipage}{1.15cm}\centering trainable\\ parameters [-] \end{minipage}\\
 & $\bar\kappa_{11}$ &$\bar\kappa_{22}$ &$\bar\kappa_{11}$ &$\bar\kappa_{22}$  &&&
  \\ \hline \hline &&&&&&&  \\[-2mm] 
  volume fraction only              &6.99\,\tiny{(5.60)}&6.86\,\tiny{(5.81)}&5.89\,\tiny{(4.14)}&5.54\,\tiny{(4.24)}&8.91\,\tiny{(7.47)}&144.12\,\tiny{(91.2)} & 241  \\%
  reference model \cite{lissner2019data} &1.92\,\tiny{(1.62)}&1.91\,\tiny{(1.63)}&1.37\,\tiny{(1.28)}&1.30\,\tiny{(1.24)}&2.39\,\tiny{(1.99)}&10.40\,\tiny{(6.47)} & 241 %
  \\  \hline &&&&&&&  \\[-2mm]
  feature iteration (1)             &1.40\,\tiny{(1.15)}&1.13\,\tiny{(1.19)}&1.04\,\tiny{(0.91)}&0.83\,\tiny{(0.96)}&1.90\,\tiny{(1.72)}&6.58\,\tiny{(4.84)} & 4\,997 \\ %
  feature iteration (2)             &1.26\,\tiny{(1.01)}&1.20\,\tiny{(1.02)}&1.00\,\tiny{(0.81)}&0.91\,\tiny{(0.79)}&1.71\,\tiny{(1.38)}&5.33\,\tiny{(3.11)} & 4\,997 %
  \\  \hline &&&&&&&  \\[-2mm]
  generic Conv Net                  &1.87\,\tiny{(1.91)}&1.81\,\tiny{(1.85)}&1.46\,\tiny{(1.55)}&1.41\,\tiny{(1.49)}&2.37\,\tiny{(2.29)}&10.21\,\tiny{(8.55)} & 220\,331 \\ %
  \begin{minipage}{2.6cm}generic Conv Net \\ \,+ vol bypass \end{minipage}                          &1.70\,\tiny{(1.56)}&1.56\,\tiny{(1.46)}&1.21\,\tiny{(1.28)}&1.14\,\tiny{(1.21)}&4.91\,\tiny{(1.98)}&43.79\,\tiny{(6.43)} & 220\,359 \\[4mm] %
  \begin{minipage}{2.6cm}inception Conv Net \\ \,+ vol bypass \end{minipage}                        &2.01\,\tiny{(1.90)}&2.11\,\tiny{(1.84)}&1.49\,\tiny{(1.51)}&1.65\,\tiny{(1.47)}&2.68\,\tiny{(2.43)}&13.07\,\tiny{(9.65)} & 278\,831 \\[4mm] %
  \begin{minipage}{2.6cm}generic Conv Net \\ \,+ vol bypass\\ \,+data augmentation \end{minipage}   &$10^8$\,\tiny{(0.88)}&$10^7$\,\tiny{(0.90)}&0.65\,\tiny{(0.72)}&0.70\,\tiny{(0.71)}&$10^8$\,\tiny{(1.23)}&$10^{17}$\,\tiny{(2.45)} & 220\,359 
   \\[4mm]  %\hline  &&&&&&&  \\[-5mm]
  \begin{minipage}{2.6cm}inception Conv Net \\ \,+ vol bypass\\ \,+data augmentation \end{minipage} &0.74\,\tiny{(0.77)}&0.79\,\tiny{(0.80)}&0.58\,\tiny{(0.62)}&0.62\,\tiny{(0.63)}&1.05\,\tiny{(1.04)}&2.00\,\tiny{(1.77)} & 278\,831 
    \\[4mm] \hline  &&&&&&&  \\[-2mm]%
  hybrid model                                                                                      &0.73\,\tiny{(0.68)}&0.74\,\tiny{(0.71)}&0.55\,\tiny{(0.54)}&0.56\,\tiny{(0.56)}&1.03\,\tiny{(0.93)}&1.91\,\tiny{(1.42)} & 283\,703  %
   \\ \hline \hline
\end{tabular} }
\end{table*}

In a final step, multiple BNNs (with the same architecture \tabref{tab.feature_architecture}) have been trained using only a subset of the available features. The BNNs used [6,9,12,...,51] features in each step and were trained in a 'best out of 5' setup, meaning that five BNN have been trained for each feature subset ordered by each feature selection method. The achieved validation losses are given in \figref{fig.n_features}. Generally, there is very little deviation seen between the different feature selection methods, and more strikingly, the loss decreases almost monotonously with respect to the number of features. Thus, all 51 features have been chosen for the subsequent hybrid neural network since all of the proposed features positively contribute to the prediction.

\subsection{Convolutional and hybrid neural networks}\label{sec.conv_net_results} 

\begin{figure*}%[bp] 
  \centering \includegraphics[height=0.2\textheight]{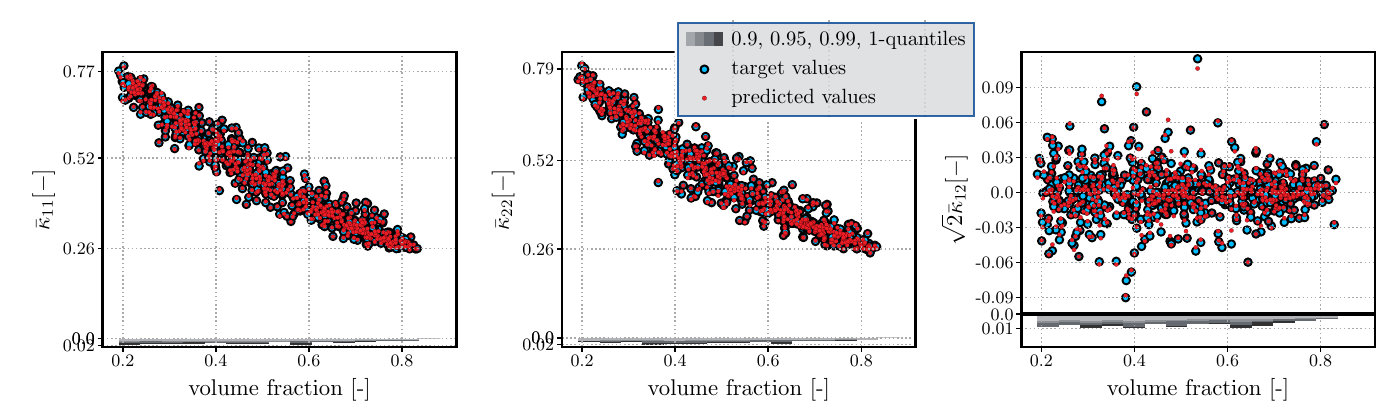}
  \caption{The plots display the target values and the respective hybrid neural network prediction for some randomly selected samples and the samples with the highest prediction error for each component of the heat conduction tensor. Each plot shares the same legend displayed in the middle. The bar plots at the bottom are computed using the absolute error and show the error quantiles on the current volume fraction interval. Note that the error bars consider the entire dataset, even the samples which were not plotted in favor of a less cluttered and better readable plot.}\label{fig.hybrid_errors}
\end{figure*}

For the convolutional neural networks (Conv Net) the aleatoric uncertainty was dropped and only one deterministic value was predicted. The deployed loss for model optimization and calibration was the mean squared error (MSE).
At first, a generic Conv Net was deployed to predict the effective heat conductivity using the available image data. The full architecture of the Conv Net is given in appendix \ref{tab.conv_architecture}. This model underperformed previous state of the art results at first, however, it has been continuously improved to outperform even the feature based approach.
Since the volume fraction is the most relevant parameter in homogenization, the plain Conv Net was improved by implementing the bypass using the volume fraction (equivalent to the hybrid model in \secref{sec.hybrid_model}), such that the Conv Net was forced to dedicate its attention in detecting high level features which quantify phenomena explaining variations around the volume fraction.
The next improvement was found by implementing the deep inception modules. The deep inception modules were both applied to the input image, and their output was flattened and concatenated into one dense regressor. The full architecture of each intermediate neueral network is given in appendix \ref{fig.inception_modules}.

The attentive reader might have noted that the loss of the model deploying deep inception modules on the benchmark set in \tabref{tab.model_errors} is slightly larger than for the generic Conv Net, however the model using deep inception modules achieved a low training loss of $1.7\cdot10^{-5}$, outperforming all previous models by a factor of 4. Thus, to fully capitalize on the deep inception modules, the dataset was enriched through the proposed data augmentation scheme, by randomly translating 50\% of the input images every 10-th epoch during training. The data augmentation scheme has also been deployed for the generic Conv Net, however leading to an accurate but strongly overfitting model, giving prohibitive prediction errors for few samples. This can be best seen in the discrepancy between the mean and median error on the test and benchmark set in \tabref{tab.model_errors}.

\begin{figure}[h] 
  \centering \includegraphics[height=0.2\textheight]{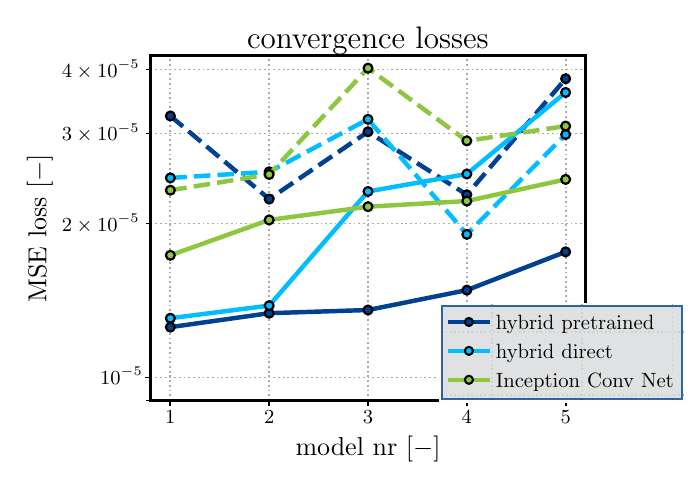} %[width=0.3\textwidth]
  \caption{The achieved validation (full lines) and training loss (dashed lines) is shown after model convergence. The hybrid neural network is compared with and without pretraining for five randomly initialized trainings, as well as the Conv Net using deep inception modules.}\label{fig.convergence}
\end{figure}

Further improvements were only found by the implementation of the hybrid neural network, adding only very few additional parameters to the inception Conv Net. As discussed in \secref{sec.hybrid_model}, the hybrid model did not always converge to a good local minimum, however, after implementing the multistage training, it reliably found an excellent minimum (\figref{fig.convergence}). Note that for almost every model the validation loss is lower than the training loss in \figref{fig.convergence}, which is explicable by the data augmentation scheme.

A more detailed error analysis of the hybrid model is given in figure format in \figref{fig.hybrid_errors}. The plot shows only a few sample predictions but it is ensures that the samples with the highest prediction errors are shown. As can be seen, the model is incredibly accurate, and is even able to accurately predict outliers of the offdiagonal component $\bar\kappa_{12}$.

\subsubsection*{Physical correctness}

\begin{figure*}%[tp]
  \centering
  \includegraphics[height=0.2\textheight]{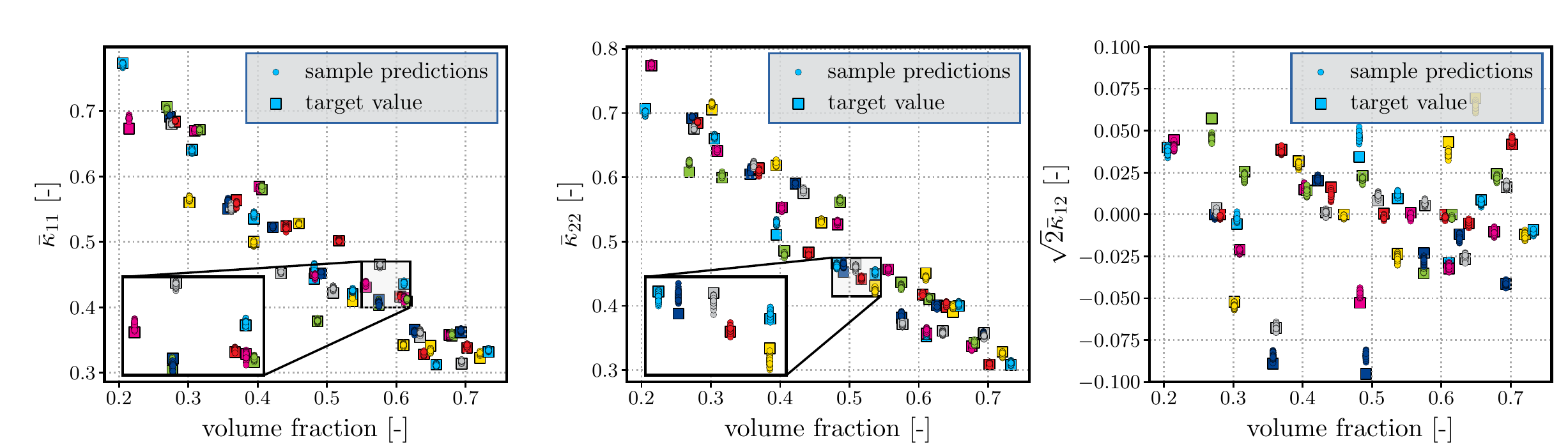}
  \caption{The prediction of the hybrid neural network is given for 50 random samples in 100 different configurations with respect to frame translation. Predictions and samples are color matched to make them distinguishable for similar volume fractions.}\label{fig.translation_invariance}
\end{figure*}

\begin{table*}%[tp]
  \caption{The error measures have been computed for the hybrid neural network once in the original configuration of the RVE and once in the rotated configuration for the entire benchmark dataset. In the rotated configuration the components $\bar{\kappa}_{11}$ and $\bar{\kappa}_{22}$ are flipped and the off-diagonal component $\bar{\kappa}_{12}$ changes its sign.  }\label{tab.rotated_errors}
\centering{\footnotesize
\begin{tabular}{|l|cc|cc|rr| } 
  \hline 
 \begin{minipage}{1.8cm}RVE image configuration \end{minipage} 
 & \multicolumn{2}{c|}{\begin{minipage}{3.0cm}\centering mean relative error [\%] \end{minipage}} 
 & \multicolumn{2}{c|}{\begin{minipage}{3.0cm}\centering median relative error [\%] \end{minipage}} 
  & \begin{minipage}{1.15cm}\centering relative\\ MSE [-] \end{minipage}
  & \begin{minipage}{1.15cm}\centering MSE \\ $10^{-5}$ [-] \end{minipage}\\
    & \begin{minipage}{1.50cm}\centering $\bar\kappa_{11}$ \end{minipage}&$\bar\kappa_{22}$ &\begin{minipage}{1.50cm}\centering $\bar\kappa_{11}$ \end{minipage}&$\bar\kappa_{22}$  &&\\
    %& $\bar\kappa_{11}$ &$\bar\kappa_{22}$ &$\bar\kappa_{11}$ &$\bar\kappa_{22}$  &&\\
      \hline &&&&&& \\[-3mm]\hline 
  original            &0.730&0.742&0.552&0.562&1.027&1.915 \\ \hline
  rotated             &0.759&0.751&0.564&0.574&1.045&1.982 \\
  \hline
\end{tabular} } 
\end{table*} 

%% translation invariance
In order to further quantify the accuracy of the hybrid neural network, a brief study regarding physical correctness is conducted. Starting out with the first property of frame translation invariance, which is not enforced but the data augmentation scheme is of assistance. Note that the previously mentioned manually derived features are designed to be frame translational invariant, with the minor exception of the volume fraction/directional edge distribution, which fluctuates ever so slightly based on the location of the pooling grid.

Some errors of the hybrid neural network with respect to frame translation are seen in \figref{fig.translation_invariance}, where it can be seen that there is no configuration of the RVE in which the model is significantly worse than in any other. The model does not yield the same prediction for every possible configuration, however, there is only slight variations between the predictions of the different configurations.

%%transposition of images
\begin{figure}[h]
  \centering \includegraphics[height=0.24\paperheight]{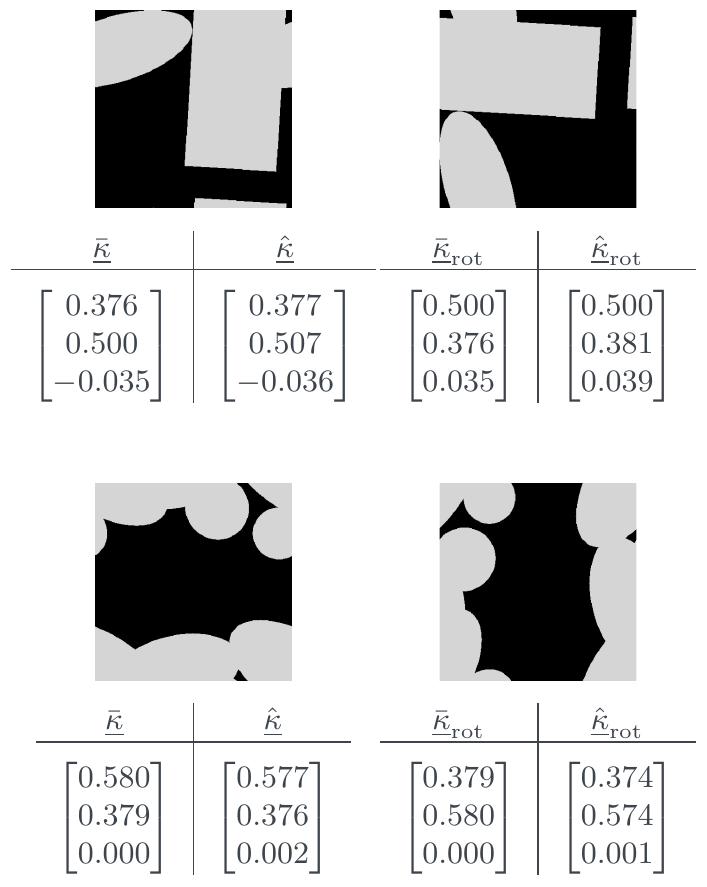}
  \caption{The prediction of two RVEs out of the benchmark dataset is given for the original representation of the RVE as well as the rotated configuration. The models prediction $\hat{\ull \kappa}$ as well as the actual target values $\bar{\ull \kappa}$ are shown below each RVE.}\label{fig.rotated_errors}
\end{figure}

%% rotation invariance
Additionally, a different geometric transform can be used to generate auxiliary samples, i.e., by rotating the RVE frames (\figref{fig.rotated_errors}). After rotation, the distinct values of the heat conduction tensor $\bar{\ul\kappa}$ are swapped and do not change, except for the sign in $\bar\kappa_{12}$. Ideally, this property should also be learned by the hybrid model, which was, similar to the translational invariance, not the case up to minor fluctuations (\tabref{tab.rotated_errors}). There seems to be an ever so slight bias in favor of the prediction of $\bar\kappa_{11}$, which implies a minor bias in the training data. The model seems to perform slightly better for the original configuration of the RVE, however, this effect is negligibly small.

Note that the rotation of the RVE frames could also be implemented for data augmentation, however the features would have to be recomputed on every augmentation step since they are, correctly so, not rotational invariant.  

The hybrid model with data augmentation is able to correctly reflect physical behaviour up to an acceptable prediction error.

\begin{table*}
  \caption{Averaged error measure over the entire benchmark dataset are shown for the model being able to predict variable phase contrast. Depending on the target value of $\bar\kappa_{\bullet}$, a relative or absolute error measure is computed, values denoted with - are not defined (c.f.\figref{fig.variable_errors}). The error measures are given at the discrete sampled phase contrasts the model was trained on.}\label{tab.variable_errors}
  \centering\footnotesize{
\begin{tabular}{|r|rr|rrr|rr| } 
  \hline
  \begin{minipage}{1.15cm} \centering Phase\\ contrast  \end{minipage}
 & \multicolumn{2}{c|}{\begin{minipage}{3.0cm}\centering mean relative error [\%] \\  $\bar\kappa_{\bullet} > 0.2$\end{minipage}} 
 & \multicolumn{3}{c|}{\begin{minipage}{3.0cm}\centering mean absolute error [-]\\ $\bar\kappa_\bullet \leq 0.2$\end{minipage}}  
   & \begin{minipage}{1.15cm}\centering relative $\sqrt{\text{MSE}}$\\ $[\%]$ \end{minipage} 
  & \begin{minipage}{1.15cm}\centering MSE\\ $\cdot 10^{-5}[-]$ \end{minipage}\\ 
    & \begin{minipage}{1.5cm}\centering$\bar\kappa_{11}$ \end{minipage}& \begin{minipage}{1.5cm}\centering$\bar\kappa_{22}$ \end{minipage}
 & $\bar\kappa_{11}$ & $\bar\kappa_{22}$ & $\bar\kappa_{12}$ 
 &
 &\\
 \hline &\begin{minipage}{1.7cm}$\quad$\end{minipage} &&&&&&\\[-2mm] %hline and spacing
\hline
2   & 0.61 & 0.59 &  -     &  -     & 0.0021 & 0.70  & 2.37\\
5   & 1.01 & 0.97 &  -     &  -     & 0.0043 & 1.33  & 3.60\\
10  & 2.07 & 2.01 & 0.0048 & 0.0053 & 0.0065 & 2.74  & 9.56\\
20  & 2.78 & 2.69 & 0.0087 & 0.0111 & 0.0090 & 4.71  & 18.98\\
50  & 3.17 & 3.36 & 0.0139 & 0.0159 & 0.0121 & 8.40  & 34.37\\
100 & 3.46 & 3.74 & 0.0159 & 0.0190 & 0.0134 & 12.16 & 44.50\\ 
\hline 
\end{tabular} } 
\end{table*}

\begin{figure*}
  \includegraphics[width=\textwidth]{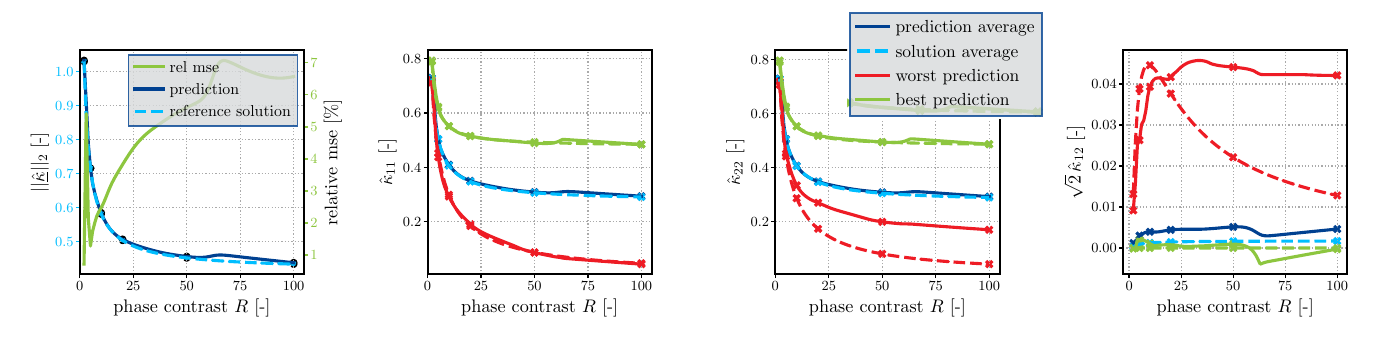}
  \caption{The prediction for variable phase contrast is given for the full interpolation range of R=[2,100] using $\Delta R=1$. To the left the averaged prediction over the entire set is give. The right three plots, which share the same legend, show the average prediction in each component, as well as the prediction of two distinct samples of the highest and lowest relative error. The predictions are always given in full lines and the reference solution in dashed lines. The phase contrasts, which the hybrid neural network has seen during training, are highlighted by dots.}\label{fig.variable_interpolation}
\end{figure*}

\subsubsection{Variable phase contrast}\label{sec.variable_contrast}

To further test the capabilities of the potent hybrid model, it was trained to predict the thermal behaviour for variable phase contrast ranging from $R=2$ to $R=100$, while considering insulating inclusions (previously, R=5 was fixed). During training, the input data of the model did not change compared to the previously discussed data, however, the output data of the model changed, where the effective heat conductivity of all training samples got computed at the discrete values of the phase contrast $R\in\{2,5,10,20,50,100\}$. In order to inform the hybrid model of the variable phase contrast, one extra input neuron was added in each branch of the model (\figref{fig.hybrid_model} the red neuron), leading to less than $500$ additional parameters. The phase contrast input parameter was linearly scaled from 0 to 1. One further adjustment was made, i.e., instead of the MSE the \relmse was used as a cost function. 
The resulting prediction errors on the benchmark set are compactly summarized in \tabref{tab.variable_errors}, and presented more elaborately in the appendix in \figref{fig.variable_errors}. There it can also be seen that the prediction becomes increasingly challenging for higher phase contrasts, which the model was still able to predict with a moderately low \relmse of 12\% for the highest phase contrast. The models accuracy for $R=5$ did slightly deteriorate compared to the previously shown hybrid model, however, in the tradeoff for generalization capabilities, where the model is able to predict every phase contrast with low prediction errors. 

One major advantage of training the different phase contrasts in a single model is the interpolation capabilities, where the prediction errors are given in \figref{fig.variable_interpolation} for $R\in{1,2,3,...,100}$. The plot shows that the smooth interpolation is in general accurately given for almost all phase contrasts within the training range with a minor exception for the short interval of $R\approx[60, 72]$, where the prediction errors are slightly larger.

\section{Summary}
The prediction of the homogenized response is improved by novel features which are developed with Bayesian assisted data mining. The aleatoric uncertainty is used to collect samples which contain characteristics inexplicable to the current feature set, commonalities within these samples were found and multiple feature descriptors were developed, efficiently quantifying these characteristics.
In addition to the manually engineered features convolutional neural networks are deployed, where we propose deep inception modules which are a priori designed to capture phenomena at different length scales within the microstructural image data. The major advantage of the deep inception modules appear to be the improved generalization capabilities and its regularizing effect. To utilize the deep inception modules to their full potential, a data augmentation scheme is presented, which can generate more than 100 000 input samples per data point, without increasing memory consumption. 

The two different neural network archetypes utilizing the newly engineered features and deep inception modules are combined into a hybrid neural network, deploying the archetypes in parallel to yield its prediction. To improve convergence behaviour, a multistage training is implemented. The resulting model was able to more than half the prediction error compared to previous state of the models. 

After extending the model to predict variable phase contrast, the proposed hybrid neural network was able to accurately predict the material response for the challenging data. It performs just slightly worse on single phase contrast compared to the model trained only on the respective data, even without touching the network layout. Remarkably, the variable contrast model is still majorly outperforming the reference model \cite{lissner2019data} by almost halving the prediction error, while enabling for completely arbitrary inputs. The model does also perform reasonably well on interpolation, yielding accurate predictions even for phase contrasts the model has not been trained on.

\section{Discussion}
One of the key motivations of machine learning is its efficient evaluation during inference. The overhead of computing all the features as well as the evaluation of the convolutional neural network is noticeable. Since all of the features, including the reduced coefficients, are computable in Fourier space or trivially obtainable, the FFT has to be conducted only once for each sample. Additionally, the convolutional kernels can be pre-computed and stored in Fourier representation when the resolution is fixed. One computational downside for the band features is that in each direction the IFFT has to be taken, since the maximum and the minimum in real space are required. Similarly to the edge distribution where the absolute value is required with its pixel location. Consequently, these features are only obtainable with an additional computational overhead. 
When timing the feature computation as well as the prediction, the model outperformed the intrinsically fast Fourier accelerated solvers by a factor of $\approx 10$, where the feature derivation made up about 80\% of the computational effort by taking $\approx 77$ seconds when considering 1500 samples, whilst the hybrid models prediction took $\approx 9$ seconds. Note that the prediction can be evaluated for multiple phase contrasts once the input feature vector is obtained. %fans 850 seconds

Further computational speedups can be obtained with additional hyperparameter tuning of the deep inception modules. However, the obtained improvement is already significant, such that additional hyperparameter tuning has been omitted in favor of more future research, especially when considering the required time investment to explore the actual infinite range of possible variations (inception module depth, width, tuning of each branch, etc.), and the possibility of nested deep inception modules. 

Regarding the prediction accuracy, the hybrid neural network which uses the manually engineered features as well as the deep inception modules was able to significantly outperform different models. Ultimately, it has failed to learn exact physical behaviour, which is not too surprising since the prediction is solely based of a neural network without any constraints. 
\backmatter

\bmhead{Acknowledgments}
Funded by Deutsche Forschungsgemeinschaft (DFG, German Research Foundation) under Germany’s Excellence Strategy - EXC 2075 – 390740016. Contributions by Felix Fritzen are funded by Deutsche Forschungsgemeinschaft (DFG, German Research Foundation) within the Heisenberg program DFG-FR2702/8 - 406068690 and DFG-FR2702/10 - 517847245. We acknowledge the support by the Stuttgart Center for Simulation Science (SimTech).

\section*{Declarations}
% Some journals require declarations to be submitted in a standardised format. Please check the Instructions for Authors of the journal to which you are submitting to see if you need to complete this section. If yes, your manuscript must contain the following sections under the heading `Declarations':

 \begin{itemize}
\item Funding\\
Funded by Deutsche Forschungsgemeinschaft (DFG, German Research Foundation) under Germany’s Excellence Strategy - EXC 2075 – 390740016. Contributions by Felix Fritzen are funded by Deutsche Forschungsgemeinschaft (DFG, German Research Foundation) within the Heisenberg program DFG-FR2702/8 - 406068690 and DFG-FR2702/10 - 517847245. We acknowledge the support by the Stuttgart Center for Simulation Science (SimTech).
\item Conflict of interest/Competing interests\\% (check journal-specific guidelines for which heading to use)\\
    The authors declare no conflict of interest
% \item Ethics approval  - Not applicable
% \item Consent to participate - Not applicable
% \item Consent for publication - Not applicable
\item Availability of data and materials\\

The data is publicly available in the data repository of the University of Stuttgart (DaRUS) \cite{darus1151}
\item Code availability \\

The code will be made publicly available in \cite{codebase} upon publication.
% \item Authors' contributions - Not applicable
\end{itemize}

% \noindent
% If any of the sections are not relevant to your manuscript, please include the heading and write `Not applicable' for that section. 

% %%===================================================%%
% %% For presentation purpose, we have included        %%
% %% \bigskip command. please ignore this.             %%
% %%===================================================%%
% \bigskip

% Springer journals and proceedings: \url{https://www.springer.com/gp/editorial-policies}

% Nature Portfolio journals: \url{https://www.nature.com/nature-research/editorial-policies}

% \textit{Scientific Reports}: \url{https://www.nature.com/srep/journal-policies/editorial-policies}

% BMC journals: \url{https://www.biomedcentral.com/getpublished/editorial-policies}
% \end{flushleft}

\begin{appendices}




%%%%%%%%% TEXT AND DATA APPENDIX %%%%%%%%%%%%%%%
\section{Data overview}\label{appendix.data}

The binary image data $A$ is represented via voxels as a discrete field, such that we have
\begin{align}
 A & \in \ffA^{n \times n}, & A_{ij} & = A( \fx_{ij}), & \fx_{ij} & \in \varOmega,
\end{align}
where the value $A_{ij}$ denote pixel-wise constant values and the data range $\ffA \subset \ffN_0$ depends on the number of discrete phases.
In general, $A$ represents image data, more specifically the image data representing the RVE. The analyzed microstructure images are given by binary voxel data, i.e., $\ffA=\{0, 1\}$, and the resolution is fixed to $400\times 400$, i.e., $i, j \in [0,1,\dots,399]$ (using \texttt{python} notation). The binary color space $\mathbb A=\{0,1\}$ is displayed as black in the matrix phase of the material (value $0$), and the light gray foreground represents the inclusions (value 1), as can be seen in \figref{fig.data_overview}. The material is exemplary for metal-ceramic or polymer-glass composite materials. 

This study focuses on the thermal behaviour of the material, i.e., on the effective heat conductivity\footnote{Equivalently, this can be interpreted as a prediction of the permeability of the material which has the same underlying mathematical structure.}. The properties of the macroscopic material are induced by the parameters of the phases in the RVE, where a phase contrast of $R=5$ is considered such that we have
\begin{align}\label{eq.contrast}
    \kappa_{\rm M} &= 1\,\frac{\rm J}{{\rm s Km}}, &
    \kappa_{\rm I} &= \frac{\kappa_{\rm M}}R\,,
\end{align}
 i.e., low-conducting inclusions are considered. Here the subscript M denotes the matrix phase, and the subscript I denotes the inclusion phase. Simulations of the heat flow conducted via FANS \citep{leuschner2018fourier} yield the symmetric effective heat conduction tensor $\tilde{\ull\kappa}$ as 
\begin{equation}
  \tilde{\ull\kappa} = \begin{bmatrix} \tilde{\kappa}_{11} & \tilde{\kappa}_{12} \\ \tilde{\kappa}_{12} & \tilde{\kappa}_{22} \\  \end{bmatrix}\,,
\end{equation}
in the 2D setting. Due to the symmetry the target values are represented in Mandel notation
\begin{equation}
  \tilde{\ul\kappa} = \begin{bmatrix}\tilde{\kappa}_{11} \\ \tilde{\kappa}_{22} \\ \sqrt{2} \tilde{\kappa}_{12} \end{bmatrix}\,.
\end{equation}
To obtain de-dimensionalized output values the heat conduction tensor is normalized by the material property of the matrix phase, i.e.,  
\begin{equation}
  \bar{\ul\kappa} = \frac{\tilde{\ul\kappa}}{\kappa_{\rm M}}\,, 
\end{equation} 
again, given in Mandel notation. The investigated data, generated by our in-house algorithms, is made publicly available \citep{darus1151}. It hosts all of the data used for the methodological developments.

\begin{figure*}%[tp]
  \centering \includegraphics[width=0.85\textwidth]{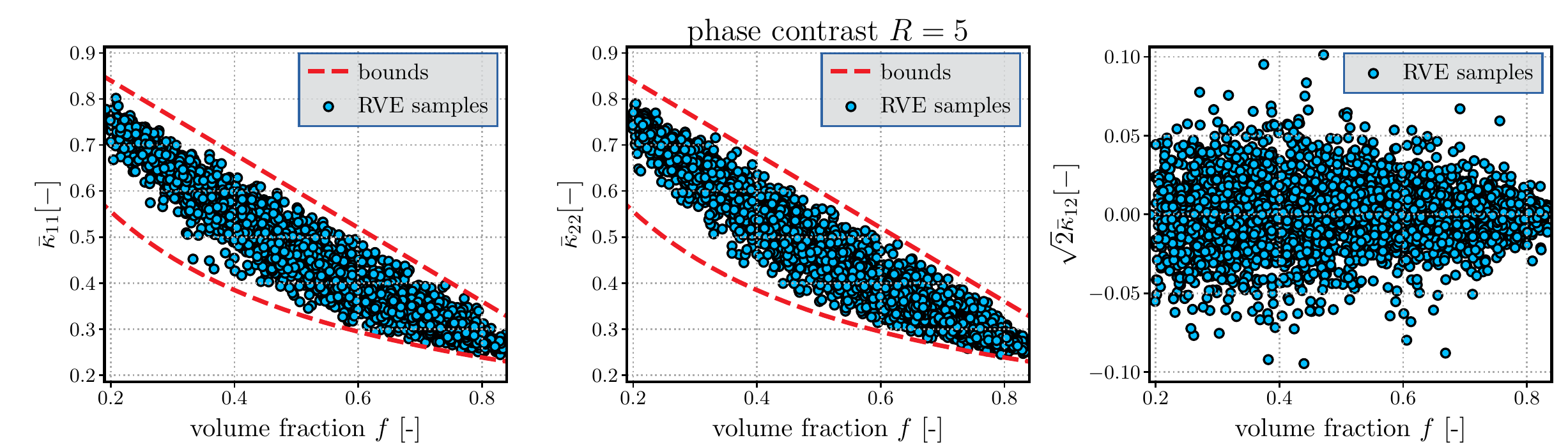}
  \caption{The effective heat conductivity for every component of $\bar{\ul\kappa}$ is related to the volume fraction for the phase contrast $R=5$. The Voigt and Reuss bounds \citep{hashin1962variational} are shown with dashed red lines for reference. The inspected data covers the admissible variation within the physical bounds pretty well.}\label{fig.kappa_scatter}
\end{figure*}

Considering the training data for the neural networks, the input data of is shifted and scaled such that each input feature of the neural network has zero mean and unit standard deviation in order to ensure statistical equivalence for all features. The transformation of  samples $X_i, i=1, \dots, n$ into $\WT{X}_i$ of a feature $X$ is computed via
\begin{align}\label{eq.scaling}
\ol{X} &= \frac{1}{n} \sum_{j=1}^n X_j, &
  \WT{X}_i  & = \frac{ \sqrt{n-1} ( X_i - \ol{X} )}{\sqrt{\displaystyle \sum_{j=1}^n ( X_j - \ol{X} )^2 }}\,.
\end{align}
This process is performed for individually for each feature, i.e., zero cross-correlation of the inputs is asserted for simplicity.
Regarding the target values $\bar{\ul\kappa}$: the diagonal components in the heat conduction tensor are well defined in $\tilde\kappa_{ii} \in (0,1)$ and the off-diagonal component $\tilde\kappa_{ij}$ fluctuates around 0 with relatively small values due to the positive definiteness of the tensor (c.f. \figref{fig.kappa_scatter}). The physically possible range of output values is consequently machine learning friendly and, therefore, no additional scaling is performed.

The complete dataset contains 30000 samples, however, only 3000 samples were used for training and validation purposes. The development of the methods were supported by using an auxiliary test set containing 1500 unseen samples. Additionally, a new set of microstructure images containing a mix of circular and rectangular inclusions within a single RVE and, further, structures composed of ellipsoidal inclusions are used as benchmark, constituting truly unseen data.

%%%%%%%%%%%%%%%%%%%%%%%%%%%%% figure appendix %%%%%%%%%%%%%%%
\section{Neural network architectures}\label{appendix.architecture}

\begin{table}[tp]
  \caption{The architecture of the Bayesian neural network used for the regression during feature selection is shown. The abbreviation BN stands for batch normalization. Every Dense connection used the selu activation function, except for the output layer which uses the identity function.}\label{tab.feature_architecture}
\centering{\scriptsize{
  \begin{tabular}{|l|cccc|}
    \hline
    &&&&\\[-2mm]
    \multirow{2}{*}{\begin{minipage}{1.5cm} \centering BNN\\ architecture \end{minipage}}&
      Dense 45 & BN & Dense 32 &BN \\
      &  Dense 25 & BN &  Dense $3\cdot2$&\\[0mm]
    \hline
\end{tabular}
}}
\end{table}

\begin{table*}%[tp]
  \caption{The chosen architecture of the generic Conv Net is shown. The convolutional layers are read as {'filter~size/stride~$\times n_{\rm channels}$'}. Dense layers simply show the number of output neurons of the layer. Every convolutional and dense layer have a selu activation functions, except for the last output layer which uses the identity function. }\label{tab.conv_architecture}
\centering{\scriptsize{
  \begin{tabular}{|l|cccc|}
    \hline
    \multirow{3}{*}{Convolutional layers}  &AvgPool 2/2& conv 11/4$\cdot32$& MaxPool 2/2& conv 7/3$\cdot$32 \\
    & MaxPool 2/2& conv 5/3$\cdot64$&   MaxPool 2/2 & conv 3/2$\cdot64$ \\
    &MaxPool 2/2& conv 3/2$\cdot96$& MaxPool 2/2& Flatten \\[2mm]
    \multirow{2}{*}{Regressor} & Dense 100&BatchNorm&Dense 70&BatchNorm \\
     & Dense 50&BatchNorm&Dense 30  &Dense 3\\
    \hline
\end{tabular}
}}
\end{table*}

\begin{figure*}%[bp]
  \centering
  \includegraphics[height=7.5cm]{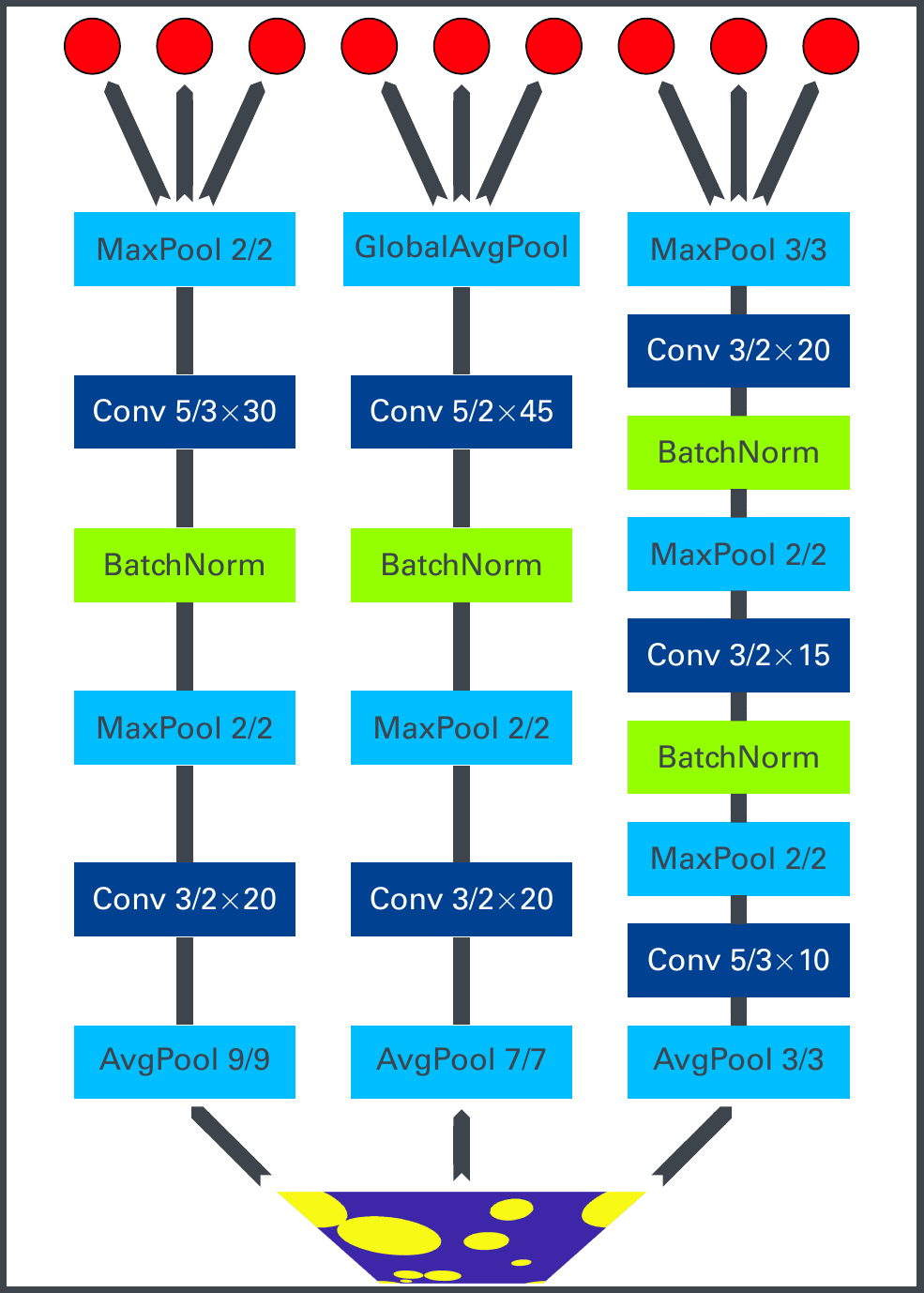}
  \includegraphics[height=7.5cm]{inception_1.pdf}
  \caption{The full layout of the deployed deep inception modules is shown. Both modules are connected in parallel to the input image, as well as the following dense regressor. These deep inception modules have been used for the final hybrid model, which contained four hidden layers of (32-32-16-16) hidden neurons using the selu activation function with batch normalization in between. }\label{fig.inception_modules}
\end{figure*}

\section{Full error plots}\label{appendix.errors}

\begin{figure*}
  \centering\includegraphics[width=0.99\textwidth]{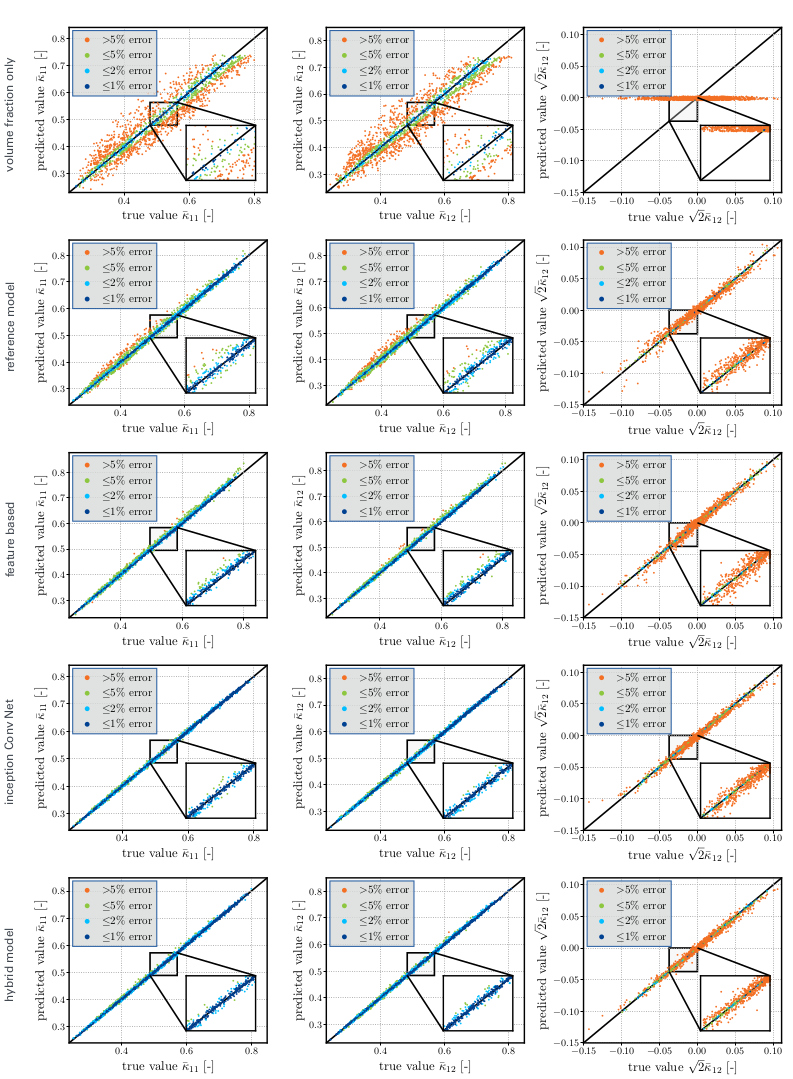}
  \caption{The R$^2$ plots for each of the relevant models is shown for the entire benchmark set and each component of the heat conduction tensor. Relative errors are color coded, such that darker scatters indicate better predictions.}\label{fig.rsquare_plots}
\end{figure*}

\begin{figure*}
  \centering\includegraphics[height=\textheight]{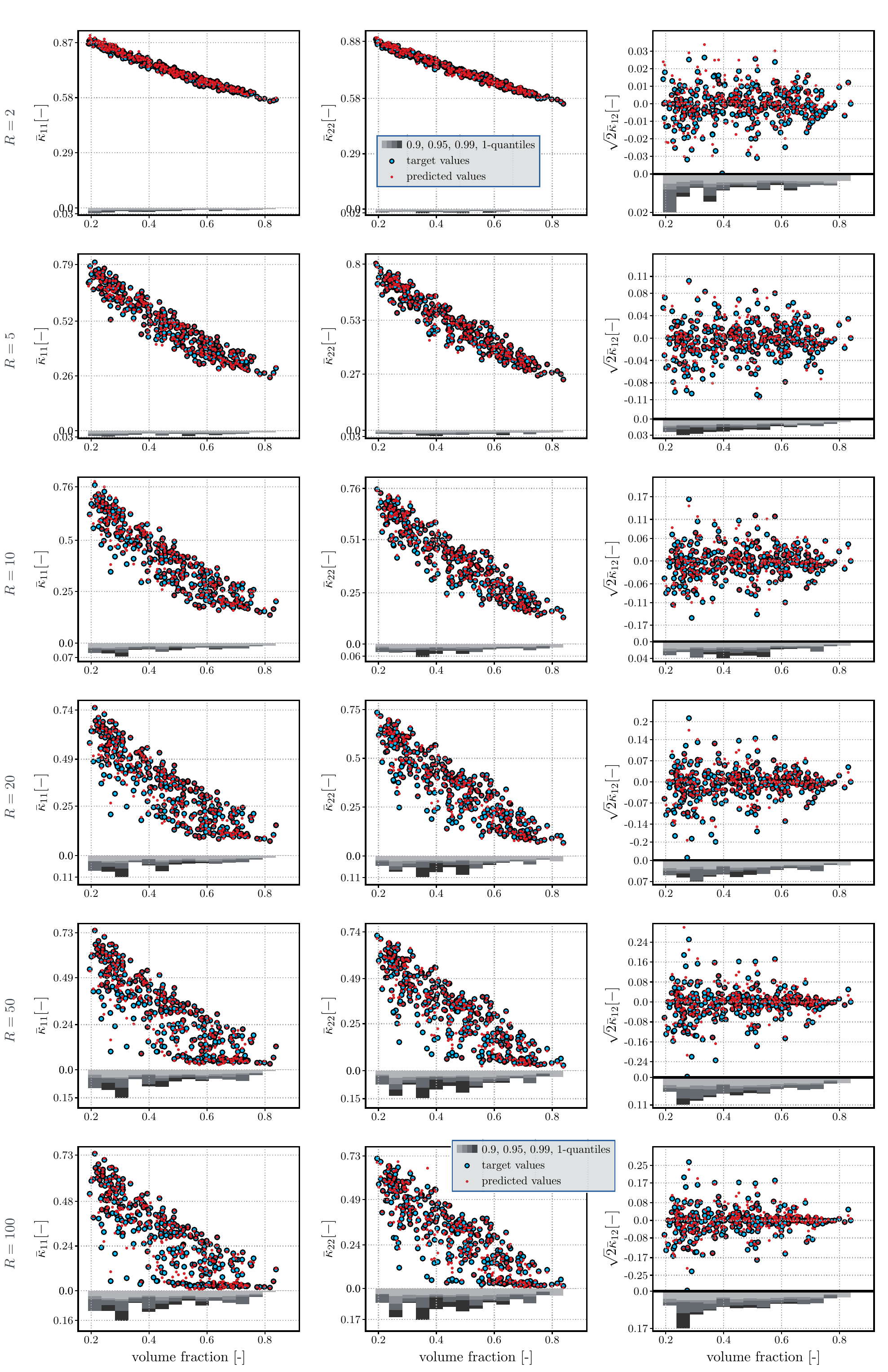}
  \caption{The predictions for all trained phase contrasts for the variable model are given alike to \figref{fig.hybrid_errors}. Not all available data points have been plotted as scatters but are considered for the error bar computation at the bottom.}\label{fig.variable_errors}
\end{figure*}

\end{appendices}


\bibliography{papers}


\begin{thebibliography}{48}
\ifx \bisbn   \undefined \def \bisbn  #1{ISBN #1}\fi
\ifx \binits  \undefined \def \binits#1{#1}\fi
\ifx \bauthor  \undefined \def \bauthor#1{#1}\fi
\ifx \batitle  \undefined \def \batitle#1{#1}\fi
\ifx \bjtitle  \undefined \def \bjtitle#1{#1}\fi
\ifx \bvolume  \undefined \def \bvolume#1{\textbf{#1}}\fi
\ifx \byear  \undefined \def \byear#1{#1}\fi
\ifx \bissue  \undefined \def \bissue#1{#1}\fi
\ifx \bfpage  \undefined \def \bfpage#1{#1}\fi
\ifx \blpage  \undefined \def \blpage #1{#1}\fi
\ifx \burl  \undefined \def \burl#1{\textsf{#1}}\fi
\ifx \doiurl  \undefined \def \doiurl#1{\url{https://doi.org/#1}}\fi
\ifx \betal  \undefined \def \betal{\textit{et al.}}\fi
\ifx \binstitute  \undefined \def \binstitute#1{#1}\fi
\ifx \binstitutionaled  \undefined \def \binstitutionaled#1{#1}\fi
\ifx \bctitle  \undefined \def \bctitle#1{#1}\fi
\ifx \beditor  \undefined \def \beditor#1{#1}\fi
\ifx \bpublisher  \undefined \def \bpublisher#1{#1}\fi
\ifx \bbtitle  \undefined \def \bbtitle#1{#1}\fi
\ifx \bedition  \undefined \def \bedition#1{#1}\fi
\ifx \bseriesno  \undefined \def \bseriesno#1{#1}\fi
\ifx \blocation  \undefined \def \blocation#1{#1}\fi
\ifx \bsertitle  \undefined \def \bsertitle#1{#1}\fi
\ifx \bsnm \undefined \def \bsnm#1{#1}\fi
\ifx \bsuffix \undefined \def \bsuffix#1{#1}\fi
\ifx \bparticle \undefined \def \bparticle#1{#1}\fi
\ifx \barticle \undefined \def \barticle#1{#1}\fi
\bibcommenthead
\ifx \bconfdate \undefined \def \bconfdate #1{#1}\fi
\ifx \botherref \undefined \def \botherref #1{#1}\fi
\ifx \url \undefined \def \url#1{\textsf{#1}}\fi
\ifx \bchapter \undefined \def \bchapter#1{#1}\fi
\ifx \bbook \undefined \def \bbook#1{#1}\fi
\ifx \bcomment \undefined \def \bcomment#1{#1}\fi
\ifx \oauthor \undefined \def \oauthor#1{#1}\fi
\ifx \citeauthoryear \undefined \def \citeauthoryear#1{#1}\fi
\ifx \endbibitem  \undefined \def \endbibitem {}\fi
\ifx \bconflocation  \undefined \def \bconflocation#1{#1}\fi
\ifx \arxivurl  \undefined \def \arxivurl#1{\textsf{#1}}\fi
\csname PreBibitemsHook\endcsname

%%% 1
\bibitem{miehe2002}
\begin{barticle}
\bauthor{\bsnm{Miehe}, \binits{C.}}:
\batitle{{Strain-driven homogenization of inelastic microstructures and
  composites based on an incremental variational formulation}}.
\bjtitle{International Journal for Numerical Methods in Engineering}
\bvolume{55},
\bfpage{1285}--\blpage{1322}
(\byear{2002}).
\doiurl{10.1002/nme.515}
\end{barticle}
\endbibitem

%%% 2
\bibitem{beyerlein2008dislocation}
\begin{barticle}
\bauthor{\bsnm{Beyerlein}, \binits{I.}},
\bauthor{\bsnm{Tom{\'e}}, \binits{C.}}:
\batitle{{A dislocation-based constitutive law for pure Zr including
  temperature effects}}.
\bjtitle{International Journal of Plasticity}
\bvolume{24}(\bissue{5}),
\bfpage{867}--\blpage{895}
(\byear{2008}).
\doiurl{10.1016/j.ijplas.2007.07.017}
\end{barticle}
\endbibitem

%%% 3
\bibitem{keshav2022fft}
\begin{botherref}
\oauthor{\bsnm{Keshav}, \binits{S.}},
\oauthor{\bsnm{Fritzen}, \binits{F.}},
\oauthor{\bsnm{Kabel}, \binits{M.}}:
{FFT-based Homogenization at Finite Strains using Composite Boxels (ComBo)}.
arXiv preprint arXiv:2204.13624
(2022)
\end{botherref}
\endbibitem

%%% 4
\bibitem{kumar2016markov}
\begin{barticle}
\bauthor{\bsnm{Kumar}, \binits{A.}},
\bauthor{\bsnm{Nguyen}, \binits{L.}},
\bauthor{\bsnm{DeGraef}, \binits{M.}},
\bauthor{\bsnm{Sundararaghavan}, \binits{V.}}:
\batitle{{A Markov random field approach for microstructure synthesis}}.
\bjtitle{Modelling and Simulation in Materials Science and Engineering}
\bvolume{24}(\bissue{3}),
\bfpage{035015}
(\byear{2016}).
\doiurl{10.1088/0965-0393/24/3/035015}
\end{barticle}
\endbibitem

%%% 5
\bibitem{cang2018improving}
\begin{barticle}
\bauthor{\bsnm{Cang}, \binits{R.}},
\bauthor{\bsnm{Li}, \binits{H.}},
\bauthor{\bsnm{Yao}, \binits{H.}},
\bauthor{\bsnm{Jiao}, \binits{Y.}},
\bauthor{\bsnm{Ren}, \binits{Y.}}:
\batitle{{Improving direct physical properties prediction of heterogeneous
  materials from imaging data via convolutional neural network and a
  morphology-aware generative model}}.
\bjtitle{Computational Materials Science}
\bvolume{150},
\bfpage{212}--\blpage{221}
(\byear{2018}).
\doiurl{10.1016/j.commatsci.2018.03.074}
\end{barticle}
\endbibitem

%%% 6
\bibitem{seibert2022descriptor}
\begin{botherref}
\oauthor{\bsnm{Seibert}, \binits{P.}},
\oauthor{\bsnm{Ra{\ss}loff}, \binits{A.}},
\oauthor{\bsnm{Ambati}, \binits{M.}},
\oauthor{\bsnm{K{\"a}stner}, \binits{M.}}:
{Descriptor-based reconstruction of three-dimensional microstructures through
  gradient-based optimization}.
Acta Materialia,
117667
(2022).
\doiurl{10.1016/j.actamat.2022.117667}
\end{botherref}
\endbibitem

%%% 7
\bibitem{lissner2019data}
\begin{barticle}
\bauthor{\bsnm{Li{\ss}ner}, \binits{J.}},
\bauthor{\bsnm{Fritzen}, \binits{F.}}:
\batitle{{Data-Driven Microstructure Property Relations}}.
\bjtitle{Mathematical and Computational Applications}
\bvolume{24}(\bissue{2}),
\bfpage{57}
(\byear{2019}).
\doiurl{10.3390/mca24020057}
\end{barticle}
\endbibitem

%%% 8
\bibitem{ford2021machine}
\begin{barticle}
\bauthor{\bsnm{Ford}, \binits{E.}},
\bauthor{\bsnm{Maneparambil}, \binits{K.}},
\bauthor{\bsnm{Rajan}, \binits{S.}},
\bauthor{\bsnm{Neithalath}, \binits{N.}}:
\batitle{{Machine learning-based accelerated property prediction of two-phase
  materials using microstructural descriptors and finite element analysis}}.
\bjtitle{Computational Materials Science}
\bvolume{191},
\bfpage{110328}
(\byear{2021}).
\doiurl{10.1016/j.commatsci.2021.110328}
\end{barticle}
\endbibitem

%%% 9
\bibitem{brough2017materials}
\begin{barticle}
\bauthor{\bsnm{Brough}, \binits{D.B.}},
\bauthor{\bsnm{Wheeler}, \binits{D.}},
\bauthor{\bsnm{Kalidindi}, \binits{S.R.}}:
\batitle{{Materials knowledge systems in python—a data science framework for
  accelerated development of hierarchical materials}}.
\bjtitle{Integrating materials and manufacturing innovation}
\bvolume{6}(\bissue{1}),
\bfpage{36}--\blpage{53}
(\byear{2017}).
\doiurl{10.1007/s40192-017-0089-0}
\end{barticle}
\endbibitem

%%% 10
\bibitem{marshall2021autonomous}
\begin{barticle}
\bauthor{\bsnm{Marshall}, \binits{A.}},
\bauthor{\bsnm{Kalidindi}, \binits{S.R.}}:
\batitle{{Autonomous development of a machine-learning model for the plastic
  response of two-phase composites from micromechanical finite element
  models}}.
\bjtitle{JOM}
\bvolume{73}(\bissue{7}),
\bfpage{2085}--\blpage{2095}
(\byear{2021}).
\doiurl{10.1007/s11837-021-04696-w}
\end{barticle}
\endbibitem

%%% 11
\bibitem{farizhandi2022processing}
\begin{botherref}
\oauthor{\bsnm{Farizhandi}, \binits{A.A.K.}},
\oauthor{\bsnm{Mamivand}, \binits{M.}}:
{Processing Time, Temperature, and Initial Chemical Composition Prediction from
  Materials Microstructure by Deep Network for Multiple Inputs and Fused Data}.
Materials \& Design,
110799
(2022)
\end{botherref}
\endbibitem

%%% 12
\bibitem{lubbers2017inferring}
\begin{barticle}
\bauthor{\bsnm{Lubbers}, \binits{N.}},
\bauthor{\bsnm{Lookman}, \binits{T.}},
\bauthor{\bsnm{Barros}, \binits{K.}}:
\batitle{{Inferring low-dimensional microstructure representations using
  convolutional neural networks}}.
\bjtitle{Physical Review E}
\bvolume{96}(\bissue{5}),
\bfpage{052111}
(\byear{2017}).
\doiurl{10.1103/PhysRevE.96.052111}
\end{barticle}
\endbibitem

%%% 13
\bibitem{gayon2020pores}
\begin{barticle}
\bauthor{\bsnm{Gayon-Lombardo}, \binits{A.}},
\bauthor{\bsnm{Mosser}, \binits{L.}},
\bauthor{\bsnm{Brandon}, \binits{N.P.}},
\bauthor{\bsnm{Cooper}, \binits{S.J.}}:
\batitle{{Pores for thought: generative adversarial networks for stochastic
  reconstruction of 3D multi-phase electrode microstructures with periodic
  boundaries}}.
\bjtitle{npj Computational Materials}
\bvolume{6}(\bissue{1}),
\bfpage{1}--\blpage{11}
(\byear{2020}).
\doiurl{10.1038/s41524-020-0340-7}
\end{barticle}
\endbibitem

%%% 14
\bibitem{liu2022correlation}
\begin{barticle}
\bauthor{\bsnm{Liu}, \binits{X.}},
\bauthor{\bsnm{Zhou}, \binits{S.}},
\bauthor{\bsnm{Yan}, \binits{Z.}},
\bauthor{\bsnm{Zhong}, \binits{Z.}},
\bauthor{\bsnm{Shikazono}, \binits{N.}},
\bauthor{\bsnm{Hara}, \binits{S.}}:
\batitle{{Correlation between microstructures and macroscopic properties of
  nickel/yttria-stabilized zirconia (Ni-YSZ) anodes: Meso-scale modeling and
  deep learning with convolutional neural networks}}.
\bjtitle{Energy and AI}
\bvolume{7},
\bfpage{100122}
(\byear{2022}).
\doiurl{10.1016/j.egyai.2021.100122}
\end{barticle}
\endbibitem

%%% 15
\bibitem{yang2020prediction}
\begin{barticle}
\bauthor{\bsnm{Yang}, \binits{C.}},
\bauthor{\bsnm{Kim}, \binits{Y.}},
\bauthor{\bsnm{Ryu}, \binits{S.}},
\bauthor{\bsnm{Gu}, \binits{G.X.}}:
\batitle{{Prediction of composite microstructure stress-strain curves using
  convolutional neural networks}}.
\bjtitle{Materials \& Design}
\bvolume{189},
\bfpage{108509}
(\byear{2020}).
\doiurl{10.1016/j.matdes.2020.108509}
\end{barticle}
\endbibitem

%%% 16
\bibitem{fernandez2020generation}
\begin{barticle}
\bauthor{\bsnm{Fern{\'a}ndez}, \binits{M.}},
\bauthor{\bsnm{Fritzen}, \binits{F.}}:
\batitle{On the generation of periodic discrete structures with identical
  two-point correlation}.
\bjtitle{Proceedings of the Royal Society A}
\bvolume{476}(\bissue{2242}),
\bfpage{20200568}
(\byear{2020})
\end{barticle}
\endbibitem

%%% 17
\bibitem{fast2011formulation}
\begin{barticle}
\bauthor{\bsnm{Fast}, \binits{T.}},
\bauthor{\bsnm{Kalidindi}, \binits{S.R.}}:
\batitle{{Formulation and calibration of higher-order elastic localization
  relationships using the MKS approach}}.
\bjtitle{Acta Materialia}
\bvolume{59}(\bissue{11}),
\bfpage{4595}--\blpage{4605}
(\byear{2011})
\end{barticle}
\endbibitem

%%% 18
\bibitem{mikut2011data}
\begin{barticle}
\bauthor{\bsnm{Mikut}, \binits{R.}},
\bauthor{\bsnm{Reischl}, \binits{M.}}:
\batitle{{Data mining tools}}.
\bjtitle{Wiley interdisciplinary reviews: data mining and knowledge discovery}
\bvolume{1}(\bissue{5}),
\bfpage{431}--\blpage{443}
(\byear{2011})
\end{barticle}
\endbibitem

%%% 19
\bibitem{tipping2003bayesian}
\begin{bchapter}
\bauthor{\bsnm{Tipping}, \binits{M.E.}}:
\bctitle{{Bayesian inference: An introduction to principles and practice in
  machine learning}}.
In: \bbtitle{Summer School on Machine Learning},
pp. \bfpage{41}--\blpage{62}
(\byear{2003}).
\doiurl{10.1007/978-3-540-28650-9_3}.
\bcomment{Springer}
\end{bchapter}
\endbibitem

%%% 20
\bibitem{szegedy2015going}
\begin{bchapter}
\bauthor{\bsnm{Szegedy}, \binits{C.}},
\bauthor{\bsnm{Liu}, \binits{W.}},
\bauthor{\bsnm{Jia}, \binits{Y.}},
\bauthor{\bsnm{Sermanet}, \binits{P.}},
\bauthor{\bsnm{Reed}, \binits{S.}},
\bauthor{\bsnm{Anguelov}, \binits{D.}},
\bauthor{\bsnm{Erhan}, \binits{D.}},
\bauthor{\bsnm{Vanhoucke}, \binits{V.}},
\bauthor{\bsnm{Rabinovich}, \binits{A.}}:
\bctitle{{Going deeper with convolutions}}.
In: \bbtitle{Proceedings of the IEEE Conference on Computer Vision and Pattern
  Recognition},
pp. \bfpage{1}--\blpage{9}
(\byear{2015})
\end{bchapter}
\endbibitem

%%% 21
\bibitem{darus1151}
\begin{botherref}
\oauthor{\bsnm{Li{\ss}ner}, \binits{J.}}:
{{2d Microstructure Data}}.
\doiurl{10.18419/darus-1151}.
\url{https://doi.org/10.18419/darus-1151}
\end{botherref}
\endbibitem

%%% 22
\bibitem{codebase}
\begin{botherref}
\oauthor{\bsnm{Li{\ss}ner}, \binits{J.}}:
Hybrid neural network codebase.
GitHub
(2023).
\url{https://github.com/J-lissner/hybrid_neural_network}
\end{botherref}
\endbibitem

%%% 23
\bibitem{torquato2013random}
\begin{bbook}
\bauthor{\bsnm{Torquato}, \binits{S.}}:
\bbtitle{Random Heterogeneous Materials: Microstructure and Macroscopic
  Properties}
vol. \bseriesno{16}.
\bpublisher{Springer}, \blocation{???}
(\byear{2013})
\end{bbook}
\endbibitem

%%% 24
\bibitem{kendall2017uncertainties}
\begin{botherref}
\oauthor{\bsnm{Kendall}, \binits{A.}},
\oauthor{\bsnm{Gal}, \binits{Y.}}:
{What uncertainties do we need in bayesian deep learning for computer vision?}
Advances in neural information processing systems
\textbf{30}
(2017)
\end{botherref}
\endbibitem

%%% 25
\bibitem{depeweg2019modeling}
\begin{botherref}
\oauthor{\bsnm{Depeweg}, \binits{S.}}:
{Modeling epistemic and aleatoric uncertainty with bayesian neural networks and
  latent variables}.
PhD thesis,
Technische Universit{\"a}t M{\"u}nchen
(2019).
\url{https://mediatum.ub.tum.de/1482483}
\end{botherref}
\endbibitem

%%% 26
\bibitem{dillon2017tensorflow}
\begin{botherref}
\oauthor{\bsnm{Dillon}, \binits{J.V.}},
\oauthor{\bsnm{Langmore}, \binits{I.}},
\oauthor{\bsnm{Tran}, \binits{D.}},
\oauthor{\bsnm{Brevdo}, \binits{E.}},
\oauthor{\bsnm{Vasudevan}, \binits{S.}},
\oauthor{\bsnm{Moore}, \binits{D.}},
\oauthor{\bsnm{Patton}, \binits{B.}},
\oauthor{\bsnm{Alemi}, \binits{A.}},
\oauthor{\bsnm{Hoffman}, \binits{M.}},
\oauthor{\bsnm{Saurous}, \binits{R.A.}}:
{Tensorflow distributions}.
arXiv preprint arXiv:1711.10604
(2017)
\end{botherref}
\endbibitem

%%% 27
\bibitem{goan2020Bayesian}
\begin{bchapter}
\bauthor{\bsnm{Goan}, \binits{E.}},
\bauthor{\bsnm{Fookes}, \binits{C.}}:
\bctitle{{Bayesian neural networks: An introduction and survey}}.
In: \bbtitle{Case Studies in Applied Bayesian Data Science},
pp. \bfpage{45}--\blpage{87}.
\bpublisher{Springer}, \blocation{???}
(\byear{2020}).
\doiurl{10.1007/978-3-030-42553-1_3}
\end{bchapter}
\endbibitem

%%% 28
\bibitem{kabir2018neural}
\begin{barticle}
\bauthor{\bsnm{Kabir}, \binits{H.D.}},
\bauthor{\bsnm{Khosravi}, \binits{A.}},
\bauthor{\bsnm{Hosen}, \binits{M.A.}},
\bauthor{\bsnm{Nahavandi}, \binits{S.}}:
\batitle{{Neural network-based uncertainty quantification: A survey of
  methodologies and applications}}.
\bjtitle{IEEE access}
\bvolume{6},
\bfpage{36218}--\blpage{36234}
(\byear{2018}).
\doiurl{10.1109/ACCESS.2018.2836917}
\end{barticle}
\endbibitem

%%% 29
\bibitem{hunt1971matrix}
\begin{barticle}
\bauthor{\bsnm{Hunt}, \binits{B.}}:
\batitle{{A matrix theory proof of the discrete convolution theorem}}.
\bjtitle{IEEE Transactions on Audio and Electroacoustics}
\bvolume{19}(\bissue{4}),
\bfpage{285}--\blpage{288}
(\byear{1971}).
\doiurl{10.1109/TAU.1971.1162202}
\end{barticle}
\endbibitem

%%% 30
\bibitem{lu1992lineal}
\begin{barticle}
\bauthor{\bsnm{Lu}, \binits{B.}},
\bauthor{\bsnm{Torquato}, \binits{S.}}:
\batitle{{Lineal-path function for random heterogeneous materials}}.
\bjtitle{Physical Review A}
\bvolume{45}(\bissue{2}),
\bfpage{922}
(\byear{1992}).
\doiurl{10.1103/PhysRevA.45.922}
\end{barticle}
\endbibitem

%%% 31
\bibitem{kalidindi2011microstructure}
\begin{barticle}
\bauthor{\bsnm{Kalidindi}, \binits{S.R.}},
\bauthor{\bsnm{Niezgoda}, \binits{S.R.}},
\bauthor{\bsnm{Salem}, \binits{A.A.}}:
\batitle{Microstructure informatics using higher-order statistics and efficient
  data-mining protocols}.
\bjtitle{Jom}
\bvolume{63}(\bissue{4}),
\bfpage{34}--\blpage{41}
(\byear{2011}).
\doiurl{10.1007/s11837-011-0057-7}
\end{barticle}
\endbibitem

%%% 32
\bibitem{scheunemann2015design}
\begin{barticle}
\bauthor{\bsnm{Scheunemann}, \binits{L.}},
\bauthor{\bsnm{Balzani}, \binits{D.}},
\bauthor{\bsnm{Brands}, \binits{D.}},
\bauthor{\bsnm{Schr{\"o}der}, \binits{J.}}:
\batitle{{Design of 3D statistically similar representative volume elements
  based on Minkowski functionals}}.
\bjtitle{Mechanics of Materials}
\bvolume{90},
\bfpage{185}--\blpage{201}
(\byear{2015}).
\doiurl{10.1016/j.mechmat.2015.03.005}
\end{barticle}
\endbibitem

%%% 33
\bibitem{balzani2014construction}
\begin{barticle}
\bauthor{\bsnm{Balzani}, \binits{D.}},
\bauthor{\bsnm{Scheunemann}, \binits{L.}},
\bauthor{\bsnm{Brands}, \binits{D.}},
\bauthor{\bsnm{Schr{\"o}der}, \binits{J.}}:
\batitle{{Construction of two-and three-dimensional statistically similar RVEs
  for coupled micro-macro simulations}}.
\bjtitle{Computational Mechanics}
\bvolume{54}(\bissue{5}),
\bfpage{1269}--\blpage{1284}
(\byear{2014})
\end{barticle}
\endbibitem

%%% 34
\bibitem{gholamalinezhad2020pooling}
\begin{botherref}
\oauthor{\bsnm{Gholamalinezhad}, \binits{H.}},
\oauthor{\bsnm{Khosravi}, \binits{H.}}:
{Pooling methods in deep neural networks, a review}.
arXiv preprint arXiv:2009.07485
(2020)
\end{botherref}
\endbibitem

%%% 35
\bibitem{vernon1991machine}
\begin{bbook}
\bauthor{\bsnm{Vernon}, \binits{D.}}:
\bbtitle{Machine Vision: Automated Visual Inspection and Robot Vision}.
\bpublisher{Prentice-Hall, Inc.}, \blocation{???}
(\byear{1991})
\end{bbook}
\endbibitem

%%% 36
\bibitem{o2015introduction}
\begin{botherref}
\oauthor{\bsnm{O'Shea}, \binits{K.}},
\oauthor{\bsnm{Nash}, \binits{R.}}:
{An introduction to convolutional neural networks}.
arXiv preprint arXiv:1511.08458
(2015)
\end{botherref}
\endbibitem

%%% 37
\bibitem{basheer2000artificial}
\begin{barticle}
\bauthor{\bsnm{Basheer}, \binits{I.A.}},
\bauthor{\bsnm{Hajmeer}, \binits{M.}}:
\batitle{Artificial neural networks: fundamentals, computing, design, and
  application}.
\bjtitle{Journal of microbiological methods}
\bvolume{43}(\bissue{1}),
\bfpage{3}--\blpage{31}
(\byear{2000}).
\doiurl{10.1016/S0167-7012(00)00201-3}
\end{barticle}
\endbibitem

%%% 38
\bibitem{lecun1989generalization}
\begin{barticle}
\bauthor{\bsnm{LeCun}, \binits{Y.}}, \betal:
\batitle{{Generalization and network design strategies}}.
\bjtitle{Connectionism in perspective}
\bvolume{19}(\bissue{143-155}),
\bfpage{18}
(\byear{1989})
\end{barticle}
\endbibitem

%%% 39
\bibitem{schubert2019circular}
\begin{bchapter}
\bauthor{\bsnm{Schubert}, \binits{S.}},
\bauthor{\bsnm{Neubert}, \binits{P.}},
\bauthor{\bsnm{P{\"o}schmann}, \binits{J.}},
\bauthor{\bsnm{Protzel}, \binits{P.}}:
\bctitle{{Circular convolutional neural networks for panoramic images and laser
  data}}.
In: \bbtitle{2019 IEEE Intelligent Vehicles Symposium (IV)},
pp. \bfpage{653}--\blpage{660}
(\byear{2019}).
\bcomment{IEEE}
\end{bchapter}
\endbibitem

%%% 40
\bibitem{kauderer2017quantifying}
\begin{botherref}
\oauthor{\bsnm{Kauderer-Abrams}, \binits{E.}}:
{Quantifying translation-invariance in convolutional neural networks}.
arXiv preprint arXiv:1801.01450
(2017)
\end{botherref}
\endbibitem

%%% 41
\bibitem{devries2017improved}
\begin{botherref}
\oauthor{\bsnm{DeVries}, \binits{T.}},
\oauthor{\bsnm{Taylor}, \binits{G.W.}}:
{Improved regularization of convolutional neural networks with cutout}.
arXiv preprint arXiv:1708.04552
(2017)
\end{botherref}
\endbibitem

%%% 42
\bibitem{loshchilov2017decoupled}
\begin{botherref}
\oauthor{\bsnm{Loshchilov}, \binits{I.}},
\oauthor{\bsnm{Hutter}, \binits{F.}}:
Decoupled weight decay regularization.
arXiv preprint arXiv:1711.05101
(2017)
\end{botherref}
\endbibitem

%%% 43
\bibitem{tensorflow2015whitepaper}
\begin{botherref}
\oauthor{\bsnm{Abadi}, \binits{M.}},
\oauthor{\bsnm{Agarwal}, \binits{A.}},
\oauthor{\bsnm{Barham}, \binits{P.}},
\oauthor{\bsnm{Brevdo}, \binits{E.}},
\oauthor{\bsnm{Chen}, \binits{Z.}},
\oauthor{\bsnm{Citro}, \binits{C.}},
\oauthor{\bsnm{Corrado}, \binits{G.S.}},
\oauthor{\bsnm{Davis}, \binits{A.}},
\oauthor{\bsnm{Dean}, \binits{J.}},
\oauthor{\bsnm{Devin}, \binits{M.}},
\oauthor{\bsnm{Ghemawat}, \binits{S.}},
\oauthor{\bsnm{Goodfellow}, \binits{I.}},
\oauthor{\bsnm{Harp}, \binits{A.}},
\oauthor{\bsnm{Irving}, \binits{G.}},
\oauthor{\bsnm{Isard}, \binits{M.}},
\oauthor{\bsnm{Yangqing}, \binits{J.}},
\oauthor{\bsnm{Jozefowicz}, \binits{R.}},
\oauthor{\bsnm{Kaiser}, \binits{L.}},
\oauthor{\bsnm{Kudlur}, \binits{M.}},
\oauthor{\bsnm{Levenberg}, \binits{J.}},
\oauthor{\bsnm{Man\'{e}}, \binits{D.}},
\oauthor{\bsnm{Monga}, \binits{R.}},
\oauthor{\bsnm{Moore}, \binits{S.}},
\oauthor{\bsnm{Murray}, \binits{D.}},
\oauthor{\bsnm{Olah}, \binits{C.}},
\oauthor{\bsnm{Schuster}, \binits{M.}},
\oauthor{\bsnm{Shlens}, \binits{J.}},
\oauthor{\bsnm{Steiner}, \binits{B.}},
\oauthor{\bsnm{Sutskever}, \binits{I.}},
\oauthor{\bsnm{Talwar}, \binits{K.}},
\oauthor{\bsnm{Tucker}, \binits{P.}},
\oauthor{\bsnm{Vanhoucke}, \binits{V.}},
\oauthor{\bsnm{Vasudevan}, \binits{V.}},
\oauthor{\bsnm{Vi\'{e}gas}, \binits{F.}},
\oauthor{\bsnm{Vinyals}, \binits{O.}},
\oauthor{\bsnm{Warden}, \binits{P.}},
\oauthor{\bsnm{Wattenberg}, \binits{M.}},
\oauthor{\bsnm{Wicke}, \binits{M.}},
\oauthor{\bsnm{Yu}, \binits{Y.}},
\oauthor{\bsnm{Zheng}, \binits{X.}}:
{ {TensorFlow}: Large-Scale Machine Learning on Heterogeneous Systems}.
Software available from tensorflow.org
(2015).
\url{https://www.tensorflow.org/}
\end{botherref}
\endbibitem

%%% 44
\bibitem{feir1974empirical}
\begin{barticle}
\bauthor{\bsnm{Feir-Walsh}, \binits{B.J.}},
\bauthor{\bsnm{Toothaker}, \binits{L.E.}}:
\batitle{{An empirical comparison of the ANOVA F-test, normal scores test and
  Kruskal-Wallis test under violation of assumptions}}.
\bjtitle{Educational and Psychological Measurement}
\bvolume{34}(\bissue{4}),
\bfpage{789}--\blpage{799}
(\byear{1974}).
\doiurl{10.1177/001316447403400406}
\end{barticle}
\endbibitem

%%% 45
\bibitem{guyon2002gene}
\begin{barticle}
\bauthor{\bsnm{Guyon}, \binits{I.}},
\bauthor{\bsnm{Weston}, \binits{J.}},
\bauthor{\bsnm{Barnhill}, \binits{S.}},
\bauthor{\bsnm{Vapnik}, \binits{V.}}:
\batitle{{Gene selection for cancer classification using support vector
  machines}}.
\bjtitle{Machine learning}
\bvolume{46}(\bissue{1}),
\bfpage{389}--\blpage{422}
(\byear{2002}).
\doiurl{10.1023/A:1012487302797}
\end{barticle}
\endbibitem

%%% 46
\bibitem{geurts2006extremely}
\begin{barticle}
\bauthor{\bsnm{Geurts}, \binits{P.}},
\bauthor{\bsnm{Ernst}, \binits{D.}},
\bauthor{\bsnm{Wehenkel}, \binits{L.}}:
\batitle{{Extremely randomized trees}}.
\bjtitle{Machine learning}
\bvolume{63}(\bissue{1}),
\bfpage{3}--\blpage{42}
(\byear{2006}).
\doiurl{10.1007/s10994-006-6226-1}
\end{barticle}
\endbibitem

%%% 47
\bibitem{leuschner2018fourier}
\begin{barticle}
\bauthor{\bsnm{Leuschner}, \binits{M.}},
\bauthor{\bsnm{Fritzen}, \binits{F.}}:
\batitle{{Fourier-accelerated nodal solvers (FANS) for homogenization
  problems}}.
\bjtitle{Computational Mechanics}
\bvolume{62}(\bissue{3}),
\bfpage{359}--\blpage{392}
(\byear{2018})
\end{barticle}
\endbibitem

%%% 48
\bibitem{hashin1962variational}
\begin{barticle}
\bauthor{\bsnm{Hashin}, \binits{Z.}},
\bauthor{\bsnm{Shtrikman}, \binits{S.}}:
\batitle{{A variational approach to the theory of the elastic behaviour of
  polycrystals}}.
\bjtitle{Journal of the Mechanics and Physics of Solids}
\bvolume{10}(\bissue{4}),
\bfpage{343}--\blpage{352}
(\byear{1962}).
\doiurl{10.1016/0022-5096(62)90005-4}
\end{barticle}
\endbibitem

\end{thebibliography}


\end{document}